\documentclass[10pt,journal,compsoc]{IEEEtran}
\usepackage{times}
\usepackage{epsfig}
\usepackage{graphicx}
\usepackage{amsmath}
\usepackage{amssymb}
\usepackage{booktabs}
\usepackage{xcolor}
\usepackage{array}
\usepackage{multirow}

\newcolumntype{M}{>{\centering\arraybackslash}m{.2\textwidth}}
\newcolumntype{C}[1]{>{\centering\let\newline\\\arraybackslash\hspace{0pt}}p{#1}}
\newcolumntype{R}[1]{>{\raggedleft\let\newline\\\arraybackslash\hspace{0pt}}p{#1}}
\newcolumntype{L}[1]{>{\raggedright\let\newline\\\arraybackslash\hspace{0pt}}p{#1}}

\definecolor{odotred}{RGB}{176,35,24}
\definecolor{odotgreen}{RGB}{78,173,91}
\definecolor{odotblue}{RGB}{45,112,186}
\definecolor{curveorange}{RGB}{233, 105, 24}
\definecolor{curveyellow}{RGB}{255, 224, 63}
\definecolor{curvegreen}{RGB}{5, 212, 97}
\definecolor{curvegray}{RGB}{220, 220, 220}
\definecolor{curveblue}{RGB}{148, 171, 215}
\def\eg{\emph{e.g.}~}
\def\ie{\emph{i.e.}~}
\def\etal{\emph{et al.}~}

\ifCLASSOPTIONcompsoc
  \usepackage[nocompress]{cite}
\else
  \usepackage{cite}
\fi
\ifCLASSINFOpdf
\else
\fi
\begin{document}
%
% paper title
% Titles are generally capitalized except for words such as a, an, and, as,
% at, but, by, for, in, nor, of, on, or, the, to and up, which are usually
% not capitalized unless they are the first or last word of the title.
% Linebreaks \\ can be used within to get better formatting as desired.
% Do not put math or special symbols in the title.
\title{DEMOS: Dynamic Environment Motion Synthesis in 3D Scenes via Local Spherical-BEV Perception}
%
%
% author names and IEEE memberships
% note positions of commas and nonbreaking spaces ( ~ ) LaTeX will not break
% a structure at a ~ so this keeps an author's name from being broken across
% two lines.
% use \thanks{} to gain access to the first footnote area
% a separate \thanks must be used for each paragraph as LaTeX2e's \thanks
% was not built to handle multiple paragraphs
%
%
%\IEEEcompsocitemizethanks is a special \thanks that produces the bulleted
% lists the Computer Society journals use for "first footnote" author
% affiliations. Use \IEEEcompsocthanksitem which works much like \item
% for each affiliation group. When not in compsoc mode,
% \IEEEcompsocitemizethanks becomes like \thanks and
% \IEEEcompsocthanksitem becomes a line break with idention. This
% facilitates dual compilation, although admittedly the differences in the
% desired content of \author between the different types of papers makes a
% one-size-fits-all approach a daunting prospect. For instance, compsoc 
% journal papers have the author affiliations above the "Manuscript
% received ..."  text while in non-compsoc journals this is reversed. Sigh.

\author{Jingyu Gong, Min Wang, Wentao Liu, Chen Qian, Zhizhong Zhang, Yuan Xie$^{\dag}$, and~Lizhuang~Ma$^{\dag}$% <-this % stops a space
\IEEEcompsocitemizethanks{\IEEEcompsocthanksitem J. Gong and L. Ma are with the Department of Computer Science and Engineering, Shanghai Jiao Tong University, Shanghai, China. L. Ma is also with the School of Computer Science and Technology, East China Normal University, Shanghai, China. 
% note need leading \protect in front of \\ to get a newline within \thanks as
% \\ is fragile and will error, could use \hfil\break instead.
E-mail: gongjingyu@sjtu.edu.cn, ma-lz@cs.sjtu.edu.cn.
\IEEEcompsocthanksitem M. Wang, W. Liu, and C. Qian are with the SenseTime Research, Shanghai, China. E-mail: \{wangmin, liuwentao, qianchen\}@sensetime.com.
\IEEEcompsocthanksitem Z. Zhang and Y. Xie are with the school of computer science and technology, East China Normal University, Shanghai, China. E-mail: \{zzzhang, yxie\}@cs.ecnu.edu.cn.
}% <-this % stops an unwanted space
\thanks{$^{\dag}$Corresponding authors.\\Manuscript received April 19, 2005; revised August 26, 2015.}}

% note the % following the last \IEEEmembership and also \thanks - 
% these prevent an unwanted space from occurring between the last author name
% and the end of the author line. i.e., if you had this:
% 
% \author{....lastname \thanks{...} \thanks{...} }
%                     ^------------^------------^----Do not want these spaces!
%
% a space would be appended to the last name and could cause every name on that
% line to be shifted left slightly. This is one of those "LaTeX things". For
% instance, "\textbf{A} \textbf{B}" will typeset as "A B" not "AB". To get
% "AB" then you have to do: "\textbf{A}\textbf{B}"
% \thanks is no different in this regard, so shield the last } of each \thanks
% that ends a line with a % and do not let a space in before the next \thanks.
% Spaces after \IEEEmembership other than the last one are OK (and needed) as
% you are supposed to have spaces between the names. For what it is worth,
% this is a minor point as most people would not even notice if the said evil
% space somehow managed to creep in.

% The paper headers
\markboth{Journal of \LaTeX\ Class Files,~Vol.~14, No.~8, August~2015}%
{Shell \MakeLowercase{\textit{et al.}}: Bare Demo of IEEEtran.cls for Computer Society Journals}
% The only time the second header will appear is for the odd numbered pages
% after the title page when using the twoside option.
% 
% *** Note that you probably will NOT want to include the author's ***
% *** name in the headers of peer review papers.                   ***
% You can use \ifCLASSOPTIONpeerreview for conditional compilation here if
% you desire.

% The publisher's ID mark at the bottom of the page is less important with
% Computer Society journal papers as those publications place the marks
% outside of the main text columns and, therefore, unlike regular IEEE
% journals, the available text space is not reduced by their presence.
% If you want to put a publisher's ID mark on the page you can do it like
% this:
%\IEEEpubid{0000--0000/00\$00.00~\copyright~2015 IEEE}
% or like this to get the Computer Society new two part style.
%\IEEEpubid{\makebox[\columnwidth]{\hfill 0000--0000/00/\$00.00~\copyright~2015 IEEE}%
%\hspace{\columnsep}\makebox[\columnwidth]{Published by the IEEE Computer Society\hfill}}
% Remember, if you use this you must call \IEEEpubidadjcol in the second
% column for its text to clear the IEEEpubid mark (Computer Society jorunal
% papers don't need this extra clearance.)

% use for special paper notices
%\IEEEspecialpapernotice{(Invited Paper)}

% for Computer Society papers, we must declare the abstract and index terms
% PRIOR to the title within the \IEEEtitleabstractindextext IEEEtran
% command as these need to go into the title area created by \maketitle.
% As a general rule, do not put math, special symbols or citations
% in the abstract or keywords.
\IEEEtitleabstractindextext{%
\begin{abstract}
Motion synthesis in real-world 3D scenes has recently attracted much attention. However, the static environment assumption made by most current methods usually cannot be satisfied especially for real-time motion synthesis in scanned point cloud scenes, if multiple dynamic objects exist, e.g., moving persons or vehicles. To handle this problem, we propose the first Dynamic Environment MOtion Synthesis framework (DEMOS) to predict future motion instantly according to the current scene, and use it to dynamically update the latent motion for final motion synthesis. Concretely, we propose a Spherical-BEV perception method to extract local scene features that are specifically designed for instant scene-aware motion prediction. Then, we design a time-variant motion blending to fuse the new predicted motions into the latent motion, and the final motion is derived from the updated latent motions, benefitting both from motion-prior and iterative methods. We unify the data format of two prevailing datasets, PROX and GTA-IM, and take them for motion synthesis evaluation in 3D scenes. We also assess the effectiveness of the proposed method in dynamic environments from GTA-IM and Semantic3D to check the responsiveness. The results show our method outperforms previous works significantly and has great performance in handling dynamic environments.
\end{abstract}

% Note that keywords are not normally used for peerreview papers.
}

% make the title area
\maketitle

% To allow for easy dual compilation without having to reenter the
% abstract/keywords data, the \IEEEtitleabstractindextext text will
% not be used in maketitle, but will appear (i.e., to be "transported")
% here as \IEEEdisplaynontitleabstractindextext when the compsoc 
% or transmag modes are not selected <OR> if conference mode is selected 
% - because all conference papers position the abstract like regular
% papers do.
\IEEEdisplaynontitleabstractindextext
% \IEEEdisplaynontitleabstractindextext has no effect when using
% compsoc or transmag under a non-conference mode.

% For peer review papers, you can put extra information on the cover
% page as needed:
% \ifCLASSOPTIONpeerreview
% \begin{center} \bfseries EDICS Category: 3-BBND \end{center}
% \fi
%
% For peerreview papers, this IEEEtran command inserts a page break and
% creates the second title. It will be ignored for other modes.
\IEEEpeerreviewmaketitle

\IEEEraisesectionheading{\section{Introduction}\label{sec:intro}}
% Computer Society journal (but not conference!) papers do something unusual
% with the very first section heading (almost always called "Introduction").
% They place it ABOVE the main text! IEEEtran.cls does not automatically do
% this for you, but you can achieve this effect with the provided
% \IEEEraisesectionheading{} command. Note the need to keep any \label that
% is to refer to the section immediately after \section in the above as
% \IEEEraisesectionheading puts \section within a raised box.

% The very first letter is a 2 line initial drop letter followed
% by the rest of the first word in caps (small caps for compsoc).
% 
% form to use if the first word consists of a single letter:
% \IEEEPARstart{A}{demo} file is ....
% 
% form to use if you need the single drop letter followed by
% normal text (unknown if ever used by the IEEE):
% \IEEEPARstart{A}{}demo file is ....
% 
% Some journals put the first two words in caps:
% \IEEEPARstart{T}{his demo} file is ....
% 
% Here we have the typical use of a "T" for an initial drop letter
% and "HIS" in caps to complete the first word.
\IEEEPARstart{S}{cene-aware} motion synthesis, in which we generate human motion sequence in real-world 3D scanned scenes, has attracted a lot of attention recently due to its wide application in robotics navigation, virtual/augmented reality, and simulated data synthesis~\cite{cao2020long,wang2021synthesizing,wang2022towards}. However, real-time motion generation in 3D scanned scene point clouds is quite challenging due to the implicit constraint on human-scene interaction and the irregular structure of scanned scene point cloud~\cite{qi2017pointnet,thomas2019kpconv,zhang2020place,guo2020deep}.

Recent works paid more attention to learning the human scene interaction based on the parametric human body model ~\cite{loper2015smpl,pavlakos2019expressive}, 3D motion and scene datasets~\cite{ionescu2013human3,armeni20163d,dai2017scannet,chang2017matterport3d,mahmood2019amass,ren2023lidar}.  
Pioneers~\cite{zhang2020place,zhang2020psi,hassan2021populating,zhao2022compositional} explored the relationship between human bodies and static scenes by utilizing various scene hints like occupation, distance, and semantics. 
Further, iterative methods~\cite{starke2019nsm,hassan2021stochastic} are used to predict future poses gradually in a simulated virtual world with explicit scene constraints. As for motion synthesis in real scanned point cloud scenes, prior-based works~\cite{wang2021synthesizing,wang2022towards} attempted to directly generate entire motion sequences rather than in an iterative manner, and took a further optimization stage to maximize the plausibility of human-scene interaction, acquiring higher stability from motion priors~\cite{zhang2021learning,xu2021exploring} when scene constraints are more implicit.

\begin{figure}
    \centering
    \includegraphics[width=\linewidth]{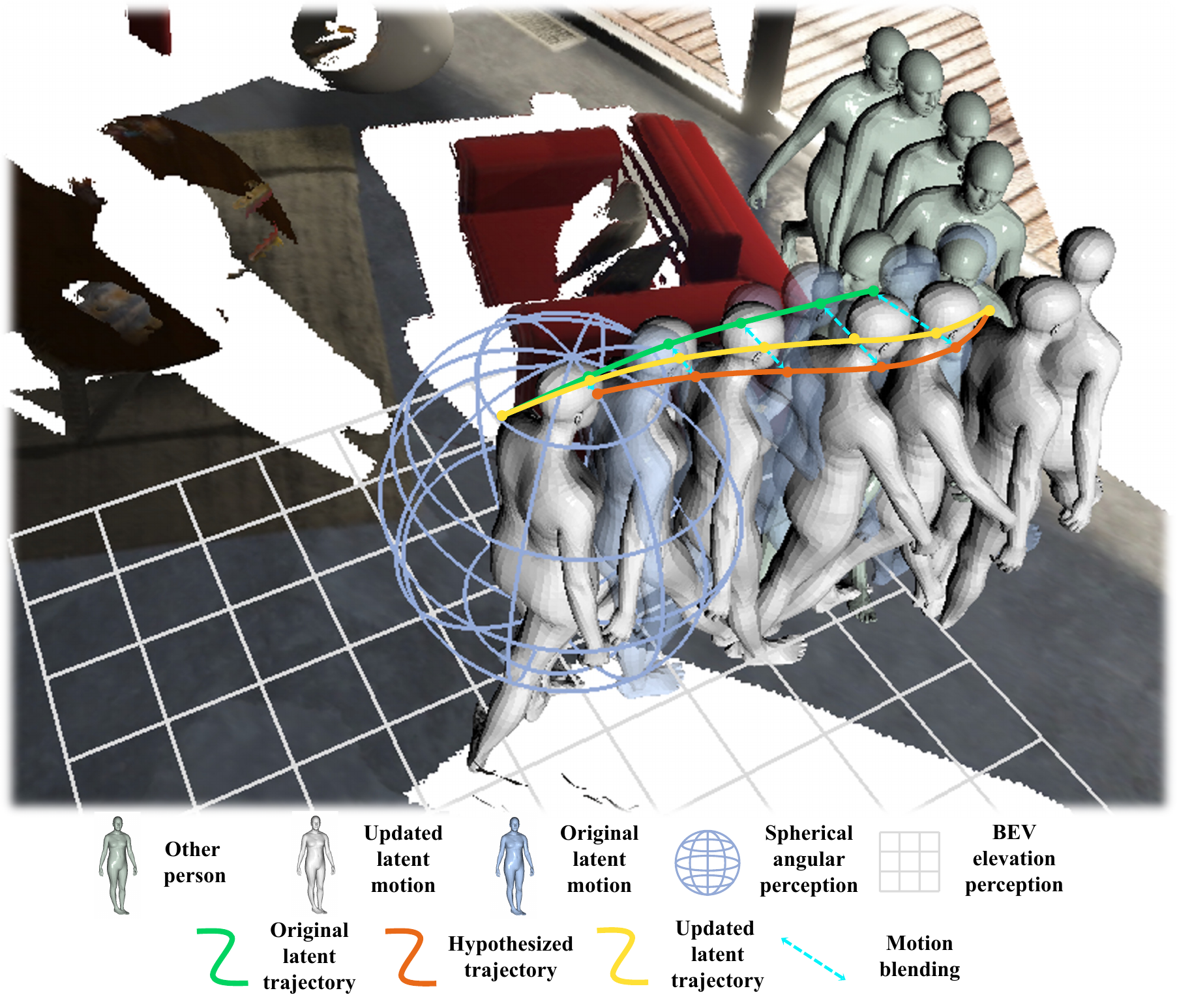}
    \caption{Illustration of proposed Dynamic Environment Motion Synthesis (DEMOS) framework based on projection-based Spherical-BEV perception. We estimate the body-centered spherical angular depth (blue spherical coordinate) and horizontal elevation map (gray mesh grid) to provide local geometry hints for instant scene-aware motion synthesis. Thus, we can iteratively generate new hypothesized motion (orange curve) and use it to update latent motion (green to yellow curve) via motion blending to adapt to the changes in scanned scene point clouds.}
    \label{fig:intro}
\end{figure}

However, they mainly considered entire motion synthesis and optimization in static scanned scenes, while in the real world, the surrounding environments may change over time. A typical situation is another person moving around as indicated in Figure~\ref{fig:intro}. This requires us to (1) predict reasonable future motion (orange curve) instantly according to the scene point clouds even without an optimization stage and (2) quickly correct and update the latent motion (green to yellow curve) to handle those changes in environments. Unfortunately, prior-based motion synthesis framework alone is unable to handle such a situation. 

In this paper, we find that, \emph{the underlying principle of a stable Dynamic Environment Motion Synthesis (DEMOS) framework for real scanned point cloud scenes is a collaboration of prior-based instant motion generation and iterative motion updating through time-variant motion blending}.

For instant scene-aware motion generation, we need to hypothesize plausible motion according to the current environment without further optimization. In contrast to whole scene feature extraction using PointNet~\cite{qi2017pointnet} like previous works~\cite{wang2021synthesizing,wang2022towards}, we focus more on body-centered local scene structures, which is more related to human motion within seconds, especially for large-scale scene point clouds. We observe that the distances between humans and scanned scenes in different directions can well tell the local scene geometry and help predict plausible motions, and the surrounding elevation map provides important hints for future motion trajectories. Thus, two projection-based local scene perception methods are specifically designed for scene-aware motion synthesis as illustrated in Figure~\ref{fig:intro}. The blue spherical coordinate indicates the Spherical Angular Perception and gray mesh grid represents the Bird's Eye View (BEV) Elevation Perception. After that, we propose a lightweight spherical convolution network to recognize the angular depth patterns and a small network for surrounding elevation feature extraction. These informative geometry hints can make the generated motion consistent with the scanned scene point cloud.

To handle dynamic environments in scanned point cloud scenes (\eg another person walks around as shown in 
Figure~\ref{fig:intro}), we attempt to both benefit from the stability of prior-based motion generation and the responsiveness of iterative methods. We iteratively generate a future motion according to the surrounding point cloud and current start/goal information. Then, inspired by trajectory blending with user control~\cite{holden2017phase}, we design a time-variant blending for the newly generated motion and previous ones to update the latent motion, as indicated by the fusion of orange and green curves in Figure~\ref{fig:intro}. Consequently, the synthesized long-term motion will be stable due to the intrinsic plausibility of motion prior, and responsive due to the iterative update of motion.

For performance evaluation, we unify the data format of PROX~\cite{hassan2019resolving} and GTA-IM~\cite{cao2020long}, and conduct experiments on these two 3D motion-scene datasets. We compare our method with cutting-edge works in motion reconstruction quality, physical and realism metrics like previous works~\cite{wang2021synthesizing,zhang2020psi}. We also present the motion synthesis in common dynamic point cloud scenes with multi-agent or moving vehicles to show the ability of our method in handling dynamic 3D environments. Comprehensive experiments show our method outperforms prevailing ones by a large margin and works well in handling dynamic environments.

Overall, our major contribution can be summarized as follows: (1) We propose the first Dynamic Environment MOtion Synthesis (DEMOS) framework to handle the real-time changes in 3D scanned point cloud scenes. (2) We design a Spherical-BEV perception method to recognize local geometry specifically for instant scene-aware motion generation. (3) We introduce motion blending as a bridge of motion-prior and iteration-based methods to adapt the synthesized motion to dynamic scanned scenes. (4) We align the data format of PROX and GTA-IM and achieve SOTA performance on motion prediction and synthesis in both static and dynamic environments.

\section{Related Work}
\label{sec:related}
\textbf{3D Scene Context Perception.} Various scene feature extraction methods~\cite{qi2017pointnet,shen2018mining,wu2019pointconv,thomas2019kpconv,lei2020spherical,gong2021omni,gong2021boundary,you2021prin,hu2021learning} were proposed to better recognize the scene geometry and semantics. Recently, these scene features were taken into consideration to synthesize scene-aware poses or motions. PLACE~\cite{zhang2020place} utilized Basis Point Sets~\cite{prokudin2019efficient} to measure the distance between sampled points and human meshes, thus could better recognize the affordance information. PSI~\cite{zhang2020psi} utilized captured depth images to provide scene structure hint. POSA~\cite{hassan2021populating} attempted to place human meshes into scenes with reasonable human-scene interactions. SAMP~\cite{hassan2021stochastic} employed the voxel representation of the target object to learn the goal position and orientation. COINS~\cite{zhao2022compositional} synthesized static human bodies in scenes given action and semantic guidance, where 3D objects are jointly exploited with human body surface points in a unified latent space. MIME~\cite{yi2023mime} reversely infer the 3D environments according to the human motions. PAAK~\cite{mullen2023placing} designed to place human animations in scanned scenes by optimizing the human-scene interactions in keyframes. PointNet~\cite{qi2017pointnet} was taken to extract the feature of the whole scene which provided external hints for scene-aware motion generation~\cite{wang2021synthesizing,wang2022towards}. 

Compared with them, we propose to learn the local geometry patterns in the spherical and BEV depth representation projected from scene point clouds, which play more important roles in motion synthesis in scanned scenes.

\noindent\textbf{Motion Synthesis.} Motion synthesis including trajectory prediction and pose generation had been widely studied for a long time under different situation~\cite{alahi2014socially,alahi2016social,holden2017phase,starke2019nsm,wang2019combining,xu2021exploring,wang2021synthesizing,petrovich2021action,parsaeifard2021learning}. Pioneer works~\cite{holden2017phase,starke2019nsm} attempted to predict pose sequences iteratively in the simulated virtual worlds. PFNN~\cite{holden2017phase} took the phase information into control network weights, and NSM~\cite{starke2019nsm} additionally utilized state information to mix expert networks, allowing automatic transition between different states. LMM~\cite{holden2020learned} replaced the key operations in Motion Matching with networks. However, the explicit scene constraints in the simulated virtual world are usually unavailable for motion synthesis in real scanned point cloud scenes, resulting in a loss of stability. 
For stable motion synthesis, HMP~\cite{xu2021exploring} learned a motion prior space from a large-scale motion dataset AMASS~\cite{mahmood2019amass}, where the entire motion can be simply sampled from a normal distribution.
Based on the motion prior, Wang \etal~\cite{wang2021synthesizing} was able to generate entire motions in real scanned point cloud scenes, pursuing better stability when scene constraints are implicit. However, further optimization was still required for motion naturalness. 
Later, motion diversity~\cite{wang2022towards} is introduced to every aspect of the pipeline. DPP~\cite{parsaeifard2021learning} used LSTM only for coarse trajectory prediction and VAE for fine-grained motion forecasting.

In this paper, we aim to synthesize motion in scanned point cloud scenes instantly but consider more general dynamic environments. Compared to methods that generate motion in the virtual world, we focus on motion synthesis in real scanned point cloud scenes where scene constraints are much more implicit. For motion synthesis in real scanned scenes, we consider more general real-time motion synthesis in dynamic scene point clouds. To handle this, we introduce local Spherical-BEV perception of scenes into instant motion prediction for higher motion plausibility and stability and design time-variant motion blending to update latent motion in an iterative manner to ensure responsiveness.

\noindent\textbf{Motion Fusion.} Motion fusion through parameter blending was commonly utilized to synthesize new motions from existing motion clips~\cite{park2002line,holden2017phase,xu2020hierarchical,xu2021exploring}.  
Park \etal~\cite{park2002line} calculated the fusion weights by estimating the similarity of example motion clips and the desired one to blend the motion clips in a continuous parameter space. PFNN~\cite{holden2017phase} blended the trajectories generated from networks and those from user control for a trade-off of naturalness and responsiveness. HSNMS~\cite{xu2020hierarchical} introduced a bi-directional interpolation scheme for retrieved short-range clips to synthesize long-term motions.  HMP~\cite{xu2021exploring} also proved it is realizable to fuse motion by interpolating the latent codes in the prior space.

Inspired by these methods, we design to iteratively blend the newly generated future motion with previous latent motion using time-variant coefficients. Thus, our synthesized motion can maintain the stability of motion-prior-based methods and responsiveness of iteration-based methods.

\begin{figure*}
    \centering
    \includegraphics[width=\linewidth]{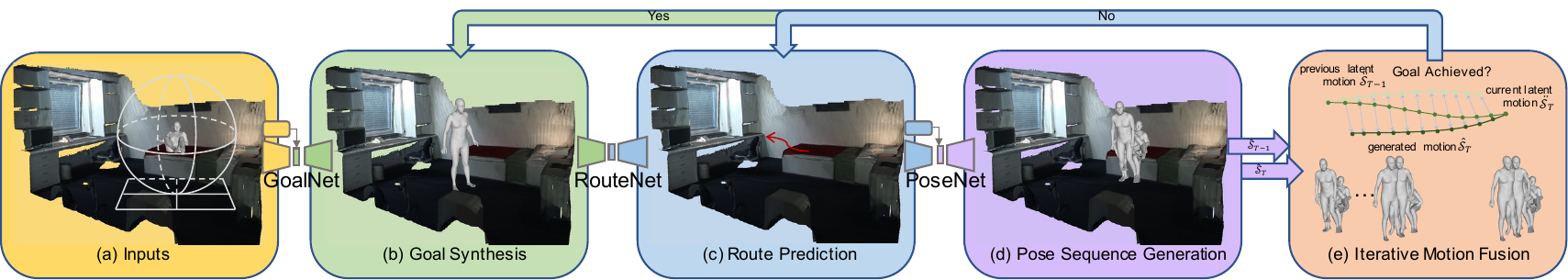}
    \caption{Framework of proposed Dynamic Environment Motion Synthesis (DEMOS) pipeline. (a) Human start information and the surrounding scene are taken as inputs for scene-aware motion synthesis. (b) We first sample goal position and orientation $\{\hat{t}_G,\hat{r}_G\}$ based on current information and the surrounding scene. (c)-(d) We then infer the future motion sequence $\hat{\mathcal{S}}_{T}$ consisting of route $\hat{\mathcal{R}}_{T}$ and pose $\hat{\mathcal{P}}_{T}$ consequently. (e) Later, the newly generated motion sequence $\hat{\mathcal{S}}_{T}$ is used to update the latent motion $\ddot{\mathcal{S}}_{T-1}$ to obtain new latent motion $\ddot{\mathcal{S}}_{T}$ in an iterative manner. The final synthesized long-term motion can be derived by $\tilde{\mathcal{S}}_{0:F}=[\ddot{\mathcal{S}}_{0}[0],\cdots,\ddot{\mathcal{S}}_{F}[0]]$.}
    \label{fig:framework}
\end{figure*}

\section{Method}
\label{sec:methods}

\subsection{Overview}

\noindent\textbf{Human Representation.} We choose the SMPL-X~\cite{pavlakos2019expressive} parameters to represent all human bodies following previous works~\cite{wang2021synthesizing,wang2022towards} to avoid irrational poses. The SMPL-X human mesh consisting of $10,475$ vertices is defined as $\mathcal{M}(t,r,\beta,\theta_{p},\theta_{h})$ where $t\in\mathbb{R}^3$ is the body root translation, $r\in\mathbb{R}^6$ is the 6D continuous rotation representation of body root orientation, $\beta\in\mathbb{R}^{10}$ is the body shape parameter, $\theta_{p}\in\mathbb{R}^{32}$ and $\theta_{h}\in\mathbb{R}^{24}$ are the compact latent codes of body pose and hand pose in prior space. Other parameters of SMPL-X are set to zeros as default for simplicity. We use point cloud to represent the scene, as it can be directly obtained from common scanning devices.

\noindent\textbf{Task Definition and DEMOS Pipeline.} In this paper, we attempt to synthesize long-term motions in dynamic scenes where other moving persons or vehicles may appear at any time in an online manner. We choose to decompose such a problem and handle this task in four steps, as shown in Figure~\ref{fig:framework}, where Spherical-BEV scene perception makes the predicted goal\&motion plausible and motion blending in a latent space makes the synthesized motion adapted to the changes in dynamic scene point clouds. The whole pipeline is formed as follows.

(a) We take human start information and scene point clond as inputs where the scene is further perceived through the Spherical-BEV projection. 

(b) GoalNet was designed to first sample the future goal position and orientation of the human body given current human and scene information. 

(c) Then, we take RouteNet to predict a scene-aware route consisting of trajectory and root orientation sequence in the short-term future. 

(d) Based on the start/goal information and route, we can generate the appropriate pose sequence according to the scene geometry using PoseNet.

(e) Later, we iteratively predict future motions and utilize newly predicted motions to update latent motion through blending for dynamic environment adaptation. 

We will finish the process or sample a new goal once the current goal is achieved. Our iterative motion synthesis method will finally generate long-term motions in dynamic environments where the surrounding point cloud may change over time.

Concretely, we estimate the future goal position distribution based on the scene structure via a CVAE GoalNet in a Bird's Eye View (BEV), and utilize the elevation map to filter out inaccessible areas. Then, we sample the goal position $\hat{t}_G$ and adjust the height according to the position elevation with state hint. We also adjust the orientation $\hat{r}_G$ according to a possible route obtained from the elevation map. 
Given current human\&scene information at time $T$ and sampled goal, we will predict the route in the future $k$ frames $\hat{\mathcal{R}}_{T}=\hat{R}_{1+T:k+T}=(\hat{t},\hat{r})_{1+T:k+T}$ according to the environment point cloud $\mathcal{E}_{T}$. We will then generate scene-aware pose sequence $\hat{\mathcal{P}}_{T}=\hat{P}_{1+T:k+T}=(\hat{\theta}_{p},\hat{\theta}_{h})_{1+T:k+T}$ which also depends on the start-goal information and predicted route. 
The motion sequence consisting of routes and poses from $T+1$ to $T+k$ is marked as $\hat{\mathcal{S}}_{T}=(\hat{\mathcal{R}},\hat{\mathcal{P}})_{T}$, termed \textbf{hypothesized motion} in this paper. 
For long-term motion synthesis, we iteratively predict $\hat{\mathcal{S}}_{T}$ and blend it with previous latent motion $\ddot{\mathcal{S}}_{T-1}$ to obtain the updated latent motion $\ddot{\mathcal{S}}_{T}$, and finally synthesize long-term motion $\tilde{\mathcal{S}}_{0:F}=[\ddot{S}_{0}[0],\cdots,\ddot{S}_{F}[0]]$ where $F$ is the total number of pre-defined frames and the \textbf{execution frame} $\ddot{S}_{i}[0]$ is the route and pose in the first frame of $\ddot{S}_{i}$. 
As for the initialization, we set $\ddot{\mathcal{S}}_{0} = \hat{\mathcal{S}}_{0}$. Here, we set $k$ to $60$ frames, \ie, $2$ seconds for a video of 30 FPS. 

\noindent\textbf{Anchor-based State Annotation.}
Human pose is highly related to the current state and the progression of state, thus previous works~\cite{starke2019nsm,hassan2021stochastic} trained the mixture of expert networks for different actions with manual state annotation. In contrast to manual annotation, we find that the human action state has a tight connection with the human-scene contact information for motion synthesis in the scanned 3D scenes. So, we annotate the state automatically according to the body part (\ie \textbf{anchors}) that is in contact with scenes without movement.

Based on the human motions from PROX~\cite{hassan2019resolving} and GTA-IM~\cite{cao2020long} datasets, we attempt to categorize all the poses into four states. For each pose from these two datasets, we first obtain the SMPL-X body mesh and check the contact~\cite{hassan2021populating} and movement~\cite{wang2021synthesizing} information of foot, gluteus, and back vertices. Thus, we can define the state of each pose according to the contact information.

\begin{figure}[htb]
    \centering
    \includegraphics[width=\linewidth]{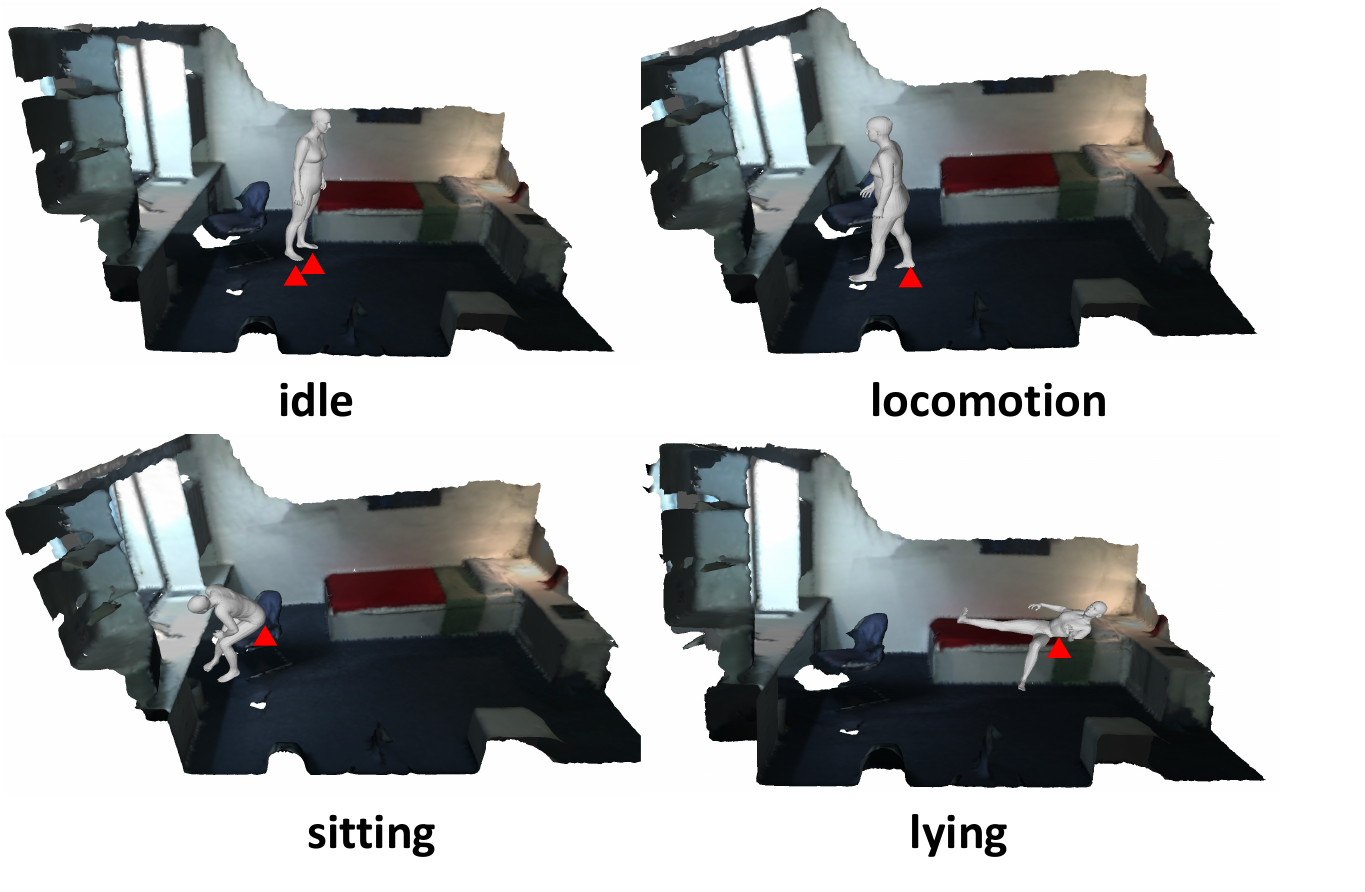}
    \caption{Illustration of states with different body parts as anchors. We annotate a pose as \textbf{idle} when two feet are anchors, \textbf{locomotion} when one foot is the anchor, \textbf{sitting} when the gluteus is the anchored part, and \textbf{lying} when the back is the anchor in our setting.}
    \label{fig:annotation}
\end{figure}

Specifically, as shown in Figure~\ref{fig:annotation}, we define bodies with only two feet as anchors to be \textbf{idle}, bodies with only one foot as the anchor to be in \textbf{locomotion}, bodies with gluteus as the anchor (and back is not an anchor) to be \textbf{sitting}, bodies with back as the anchor to be \textbf{lying}. Poses will be invalid and not be used for training/testing if none of the foot, gluteus, and back are anchored body parts.
To leverage the prior knowledge of periodic motion like locomotion~\cite{wandt20163d}, we will also automatically calculate the phase based on the foot anchor information, while the phases in other states are set to zero. Thanks to these annotations, we can utilize them to guide the scene-aware PoseNet to predict states\&phases that correspond to local scenes, and later rational poses. We can also know the anchored parts given the predicted states, which can also be used to avoid skating at contacted parts between consequent poses.

\subsection{Projection-based Local Scene Perception}
\label{subsec:perception}

\begin{figure}
    \centering
    \includegraphics[width=\linewidth]{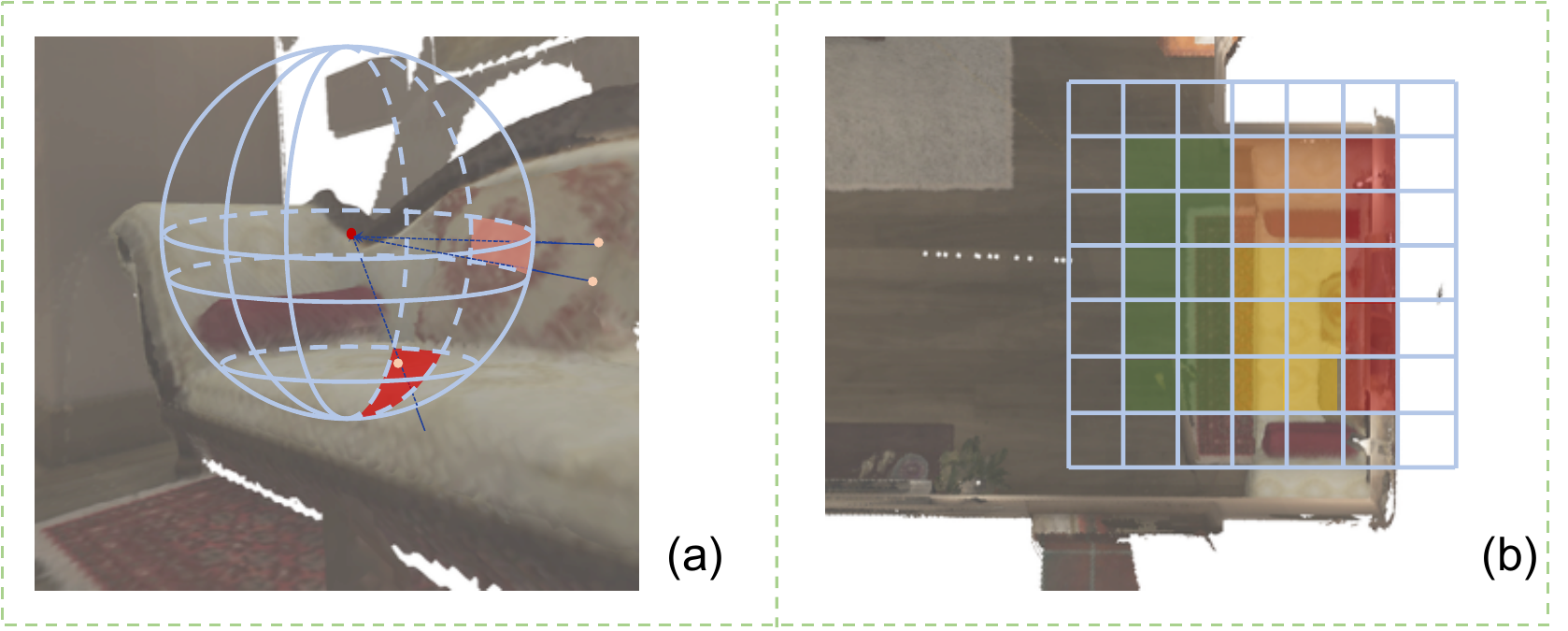}
    \caption{Illustration of Spherical Angular and BEV Elevation Perception. (a) Surrounding scene points are projected on a unit sphere to estimate the depth/distance (redder means closer) to the scene point cloud at different angular directions. (b) We take a BEV centering at the human body to recognize the surrounding elevation information (red indicates high elevation while green means low elevation).}
    \label{fig:projection}
\end{figure}

To make the synthesized motion more consistent with the surroundings, we propose two projection-based local scene perception (PBLSP) methods to extract the spherical depth and BEV elevation information from environment point clouds.

\noindent\textbf{Spherical Angular Perception.} To better understand the geometry of the surrounding scene from the human body perspective, we propose to recognize the distance to scene points in different directions from the body mesh root, as shown in Figure~\ref{fig:projection} (a). This is attributed to the observation that the state, contact information, and moving direction of human bodies have a strong relationship with the angular depth distribution pattern.

For spherical angular perception, we first extract the local scene point clouds $\mathcal{E}_T^l$ consisting of $K_1$ nearest points in environment point cloud $\mathcal{E}_T$ from human root position. Then, these $K$ points are projected onto a unit sphere with pre-defined $L_1$ longitude bins and $L_2$ latitude bins to obtain the angular depth representation of the local scene. In this paper, we simply take $L_1=36$ longitude bins and $L_2=18$ latitude bins, and $K_1=2500$ nearest points from the local scene are selected for the angular projection. This real-time point cloud projection is implemented in parallel by extending the Pytorch implementation of z-buffer~\cite{strasser1974schnelle,gong2022optde}. Specifically, we calculate the longitude and latitude bin for each point and then scatter the distance value into the sparse matrix $\mathbf{M}_1\in\mathbb{R}^{L_1\times L_2\times K_1}$ with default value $\infty$. Then, the angular depth at different directions can be easily obtained through finding the minimum among $K_1$.

After that, we choose to design a lightweight spherical convolution rather than sophisticated ones like KTN~\cite{su2019kernel} (the comparison is available in Section~\ref{subsec:ablation}) to extract the angular depth distribution pattern from the local scene point cloud. Each layer of our spherical convolution consists of a longitude convolution and a latitude convolution. Unlike the common 2D convolution, the longitude convolution with a ring structure should be cyclic. So, we first warp the feature map horizontally and apply 1D convolution on the longitude. As for the latitude convolution without ring structure, we can directly utilize 1D convolution. After a few layers of spherical convolution and angular pooling, we utilize several fully connected layers to extract the angular depth distribution feature $f_a$.

\noindent\textbf{BEV Elevation Perception.} We observe that the future trajectory and pose are highly associated with the local elevation map, as illustrated in Figure~\ref{fig:projection} (b). Take locomotion for instance, the vertical position at different frames should correspond to the elevation map to ensure foot contact. As for sitting, the seating areas in the scenes are usually typically designed with specific elevations. These scene hints can aid the proposed method in judging feasible trajectories and generating scene-aware poses.

\begin{figure*}[htb]
    \centering
    \includegraphics[width=\linewidth]{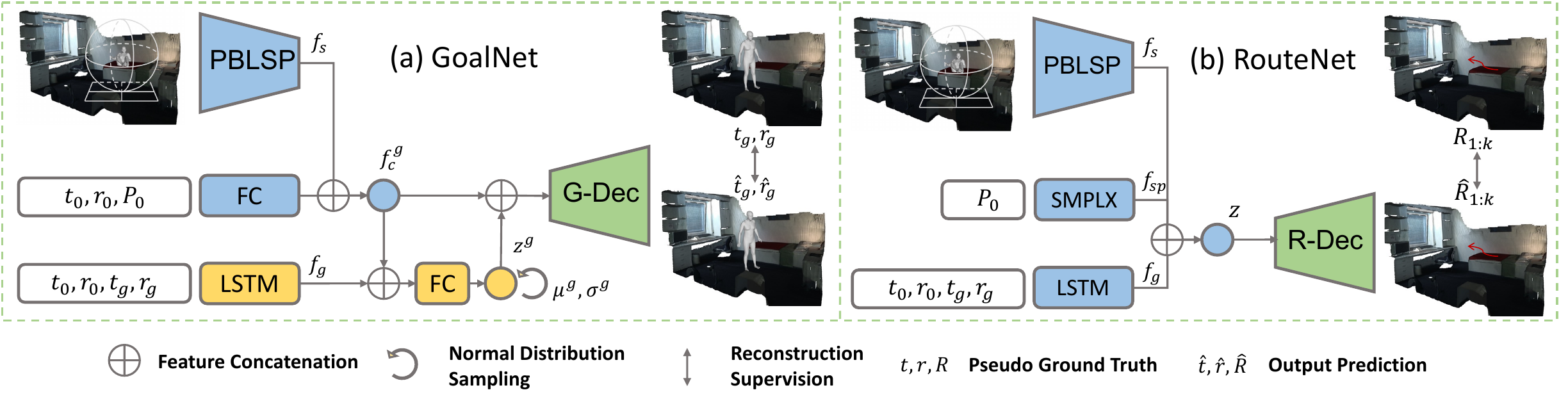}
    \caption{Frameworks of GoalNet and RouteNet. (a) GoalNet takes a CVAE structure to predict the goal $\{\hat{t}_g, \hat{r}_g\}$ based on current human body information and scene point cloud. (b) RouteNet predicts the future route $\hat{R}_{1:k}$ given the scanned scene, human start pose, and start-goal position/orientation information.}
    \label{fig:goal_route}
\end{figure*}

Thus, we attempt to take a Bird's Eye View (BEV) to extract the elevation map surrounding the current human body. In our setting, we mainly care about the elevation map of the local scene within a body-centered square with side length $S$. This square is also discretized into $M\times M$ bins, and $K_2$ surrounding points from the scene are selected for BEV depth rendering. In this paper, the side length is commonly set to $8.2$ meters which is further split into $M=41$ bins, and $K_2=20,000$ nearest points are taken for BEV perception. Different from point projection into unit sphere, there will be some points outside all the bins, and these points can not be ignored in batch operation. So, we pad the square map with one bin on each side, \ie, $L_3=M+2$ bins in total (see the bins in Figure~\ref{fig:projection} (b) without color), and all residual points are rendered to the nearest padding bins. By now, all rendered points can be scattered to another sparse matrix $\mathbf{M}_2\in\mathbb{R}^{L_3\times L_3\times K_2}$, and then we can obtain the BEV depth map $\mathcal{D}\in\mathbb{R}^{M\times M}$ by calculating the minimum depth value among $K_2$ and discarding the padding bins. After that, the BEV elevation map can be extracted from $\mathcal{B} = C_0 - \mathcal{D}$ where $C_0$ is a constant hyper-parameter and set to $0.8$ in this paper. Finally, we build a small network consisting of several 2D convolution and pooling layers to extract the BEV elevation features $f_b$.

By now, we can concatenate the scene points feature $f_p$ extracted from MLPs with the spherical angular pattern and BEV elevation feature to obtain the scene feature $f_s=f_p \oplus f_a \oplus f_b$. Such Spherical-BEV perception can supply the local scene cues essential for instant scene-aware motion synthesis given any input scenes (demonstrated in Section~\ref{subsec:ablation}). 

\subsection{Networks}

The fundamental architectures of the networks bear resemblance to those employed in previous works~\cite{wang2021synthesizing,wang2022towards} for motion synthesis. In this work, our networks focus more on how to model the correlation between human embedding and scene hints extracted from the PBLSP, and later attempt to sample feasible goals in the scene and infer rational future motion sequence under the guidance of designed anchor-based states with human-scene contact information.

\noindent\textbf{GoalNet.} Our GoalNet take a CVAE~\cite{sohn2015learning} structure shown in Figure~\ref{fig:goal_route} (a) to model the goal distribution given the current human body and scene. We concatenate the scene feature $f_s$ from the PBLSP module with the human start information to exploit the human-scene relation and obtain the goal condition feature $f_c^g$. 
Next, we extract the start-goal moving feature $f_g$ from $\{R_0,R_g\}$ and combine it with $f_c^g$ to predict the mean $\mu^g$ and variance $(\sigma^g)^2$ for the normal distribution of the goal latent code. We then sample the goal latent code $z^g$ from this distribution and predict the most likely goal $\hat{R}_g=(\hat{t}_g,\hat{r}_g)$ given the condition feature $f_c^g$ with human-scene hints.
 
For training of GoalNet, we pursue reconstructing the goal position and orientation and pushing the goal latent space close to standard normal distribution, and the supervision is formulated as:
\begin{equation}
    \mathcal{L}_{goal} = \vert\vert t_g - \hat{t}_g\vert\vert_1+\vert\vert r_g-\hat{r}_g\vert\vert_1+\lambda_1\mathcal{L}_{KL}^g,
\end{equation}
\begin{equation}
    \label{eq:klg}
    \mathcal{L}_{KL}^g = KL(\mathcal{N}(\mu^g,(\sigma^g)^2)\vert\vert\mathcal{N}(\mathbf{0},\mathbf{I})),
\end{equation}
where $KL$ represents the KL divergence between these two normal distributions, and thus $\mathcal{L}_{KL}^g$ can be simplified as:
\begin{equation}
    \label{eq:klg2}
    \mathcal{L}_{KL}^g=0.5((\sigma^g)^2+(\mu^g)^2-1-2log\sigma^g).
\end{equation}

In the motion synthesis stage, we also utilize the elevation map to filter out impossible areas in BEV, which works as a filter $\mathcal{F}$. Specifically, we build a graph based on the elevation map with each node representing a horizontal position and recording the elevation. Given the current state and position node, we can easily judge the state transition according to the relative elevation between neighboring nodes. For instance, the transition between idle to sit is usually accompanied by an elevation transition of 0.3-0.6 meters. We can then extract the possible area $\mathcal{F}$ according to the graph nodes with a possible state.

We then sample the final root position from the distribution given by a multiplication of the normal distribution $\mathcal{N}(\hat{t}_g,(\sigma^t)^2\mathbf{I})$ and the filter $\mathcal{F}$, and $\sigma^t$ is a hyper-parameter that is commonly set to $2$ in our setting. Finally, we adjust the z-axis position according to the elevation map to obtain $\hat{t}_{G}$ and adjust z-axis root orientation according to a possible path from $t_0$ to $\hat{t}_{G}$ to obtain $\hat{r}_G$.

\noindent\textbf{RouteNet.} RouteNet is designed to predict the future trajectories and orientations where surrounding scene point cloud $\mathcal{E}$, start-goal route $\{R_0,R_g\}$, and initial pose $P_0$ are taken as network inputs. The structure of RouteNet is presented in Figure~\ref{fig:goal_route} (b). As shown, we again extract the scene feature $f_s$ using the PBLSP module and take a LSTM~\cite{hochreiter1997long} block to extract the start-goal moving feature $f_g$. Meanwhile, we take the SMPL-X parameters $f_{sp}$ to represent human start pose $P_0$. Then, these three features are concatenated together to exploit the human-scene correlation and predict the route $\hat{R}_{1:k}$ in the future $k$ frames after several MLPs. To train the RouteNet, we take the L1 loss on position and orientation estimation to guide the route prediction:
\begin{equation}
    \mathcal{L}_{route} = \sum_{i=1}^k\vert\vert t_i-\hat{t}_i\vert\vert_1 + \vert\vert r_i-\hat{r}_i\vert\vert_1.
\end{equation}

In the inference stage, RouteNet can directly predict the future route within $k$ frames based on current human-scene information and sampled goal.

\begin{figure*}[htb]
    \centering
    \includegraphics[width=0.7\linewidth]{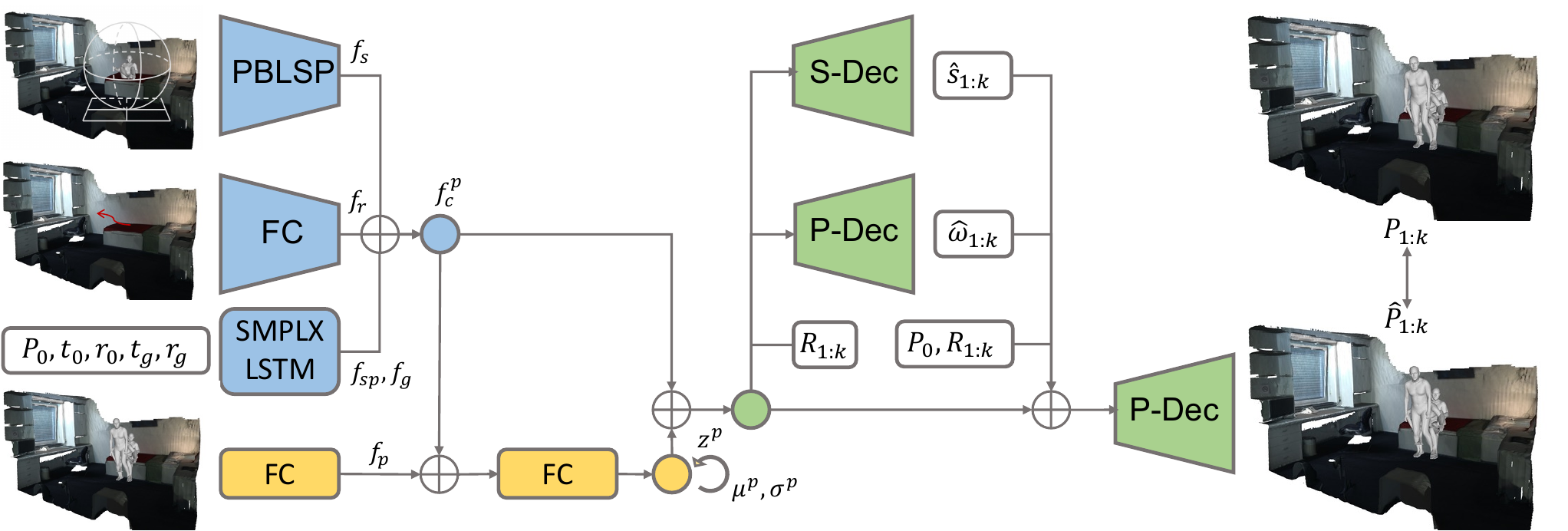}
    \caption{Framework of the proposed PoseNet which takes the form of CVAE. In the decoder, we predict the anchor-based state and phase to aid scene-aware pose sequence generation.}
    \label{fig:posenet}
\end{figure*}

\noindent\textbf{PoseNet.} As shown in Figure~\ref{fig:posenet}, we also decide to take a CVAE network structure for PoseNet to infer the future pose sequences given the current human-scene information and predicted route. For the condition encoding, we extract scene feature $f_s$, start-goal moving feature $f_g$ and start pose feature $f_{sp}$ similarly, and take a fully-connected network to encode the route feature $f_r$. These features are concatenated to obtain the pose condition feature $f_c^p = f_s\oplus f_g \oplus f_{sp} \oplus f_r$. Meanwhile, we utilize several fully connected layers to extract the pose feature $f_p$ from pose sequences $P_{1:k}$. Then, we take several MLPs to exploit the correlation of pose condition feature $f_c^p$ and pose feature $f_p$, and later extract the mean $\mu^p$ and variance $(\sigma^p)^2$ values for the normal distribution of pose latent code. After that, we can sample the pose latent vector $z^p$ from $\mathcal{N}(\mu^p,(\sigma^p)^2)$. 

Along with latent code $z^p$ and human-scene condition $f_c^p$, we choose to take an additional LSTM to extract frame-wise route features as position query, and then predict the state probability $\hat{s}_{1:k}$ at each frame accordingly. As for locomotion, we choose the relative transformations between consequent routes rather than routes themselves as input to predict the phase difference $\hat{\omega}_{1:k}$, and infer the phase sequence $\hat{p}_{1:k}$ via cumulative summation. 

Later, we use a mixture of expert networks to infer the pose sequence following previous works~\cite{starke2019nsm,hassan2021stochastic}. The mixed network weight $\alpha_i$ at each frame can be controlled by the predicted probability $\hat{s}_{1:k}$ of state $c$ at frame $i$ via $\alpha_i = \sum_{c=1}^C \hat{s}^c_i\alpha^c.$
Condition feature $f_c^p$, latent code $z^p$, initial pose $P_0$, route $R_{1:k}$, and inferred phase $\hat{p}_{1:k}$ are fed into the mixed network to predict the entire pose sequence $\hat{P}_{1:k}$.

For the training of PoseNet, we supervise the network by anticipating the intermediate state\&phase and reconstructing the input pose sequence, where loss can be formulated as follows:
\begin{equation}
    \begin{array}{rcl}
    \mathcal{L}_{pose}&=&\lambda_{2}\mathcal{L}_{KL}^p+\lambda_{3}\mathcal{L}_{joint}+\lambda_{4}\mathcal{L}_{vert}\\
    & & +\lambda_{5}\mathcal{L}_{\theta}+ \lambda_{6}\mathcal{L}_{state}+\lambda_{7}\mathcal{L}_{phase}
    \end{array}
\end{equation}
where $\mathcal{L}_{KL}^p$ is KL Divergence for pose latent code distribution. $\mathcal{L}_{joint}$, $\mathcal{L}_{vert}$ and $\mathcal{L}_{\theta}$ are the L1 loss for body joint, mesh and parameter reconstruction. $\mathcal{L}_{state}$ is the cross entropy loss in state classification and $\mathcal{L}_{phase}$ is the L1 loss in phase step prediction. 
The explicit formulation of these losses can be defined as:
\begin{center}
\begin{align}
    \mathcal{L}_{KL}^p=0.5((\sigma^p)^2+&(\mu^p)^2-1-2log\sigma^p)\label{eq:klp},\\
    \mathcal{L}_{joint} = \sum_{i=1}^k \vert\vert\mathcal{J}(t_i,r_i,\beta,&\mathcal{P}_i)-\mathcal{J}(t_i,r_i,\beta,\hat{\mathcal{P}}_i)\vert\vert_1\label{eq:node},\\
    \mathcal{L}_{vert} = \sum_{i=1}^k \vert\vert\mathcal{M}(t_i,r_i,\beta,&\mathcal{P}_i)-\mathcal{M}(t_i,r_i,\beta,\hat{\mathcal{P}}_i)\vert\vert_1\label{eq:vert},\\
    \mathcal{L}_{\theta} = \sum_{i=1}^k\vert\vert &\mathcal{P}_i - \hat{\mathcal{P}}_i\vert\vert_{1}\label{eq:theta},
\end{align}
\end{center}
\begin{center}
\begin{align}
    \mathcal{L}_{state}=\sum_{i=1}^k&\sum_{j=1}^c s_i^jlog\hat{s}_i^j\label{eq:state},\\
    \mathcal{L}_{phase}=\sum_{i=1}^k&\vert\vert \omega_i - \hat{\omega}_i\vert\vert_1.
\end{align}
\end{center}
Here, $\mu^p$ and $(\sigma^p)^2$ are the predicted mean and variance value of the latent code $z^p$ in PoseNet. $\mathcal{J}$ and $\mathcal{M}$ indicate the SMPL-X~\cite{pavlakos2019expressive} body joints and mesh vertices formed by the translation, orientation, and body shape and pose. $\hat{\mathcal{P}}_i$/$\mathcal{P}_i$ contains the predicted/P-GT body pose and hand pose parameter $(\hat{\theta}_p,\hat{\theta}_h)_i$/$(\theta_p,\theta_h)_i$. $\hat{s}_i^c$/$s_i^c$ is the predicted/P-GT probability of state $c$, while $\hat{\omega}_i$/$\omega_i$ is the predicted/P-GT phase step.

\subsection{Iterative Latent Motion Update} 
Given current human-scene information and goal, we can predict the motion in future $k$ frames, which is reasonable for a static scene. However, in dynamic scene point clouds where other characters or vehicles move around, the future motion should be updated accordingly.  So, we choose to iteratively predict future motion $\hat{\mathcal{S}}_{T}$ and use it to update previous latent motion $\ddot{\mathcal{S}}_{T-1}$ (derived from $\hat{\mathcal{S}}_{0}$, $\cdots$, $\hat{\mathcal{S}}_{T-1}$) to both benefit from the stability of prior-based motion synthesis and responsiveness of iterative methods when handling dynamic 3D environments.

Here, we introduce a time-variant motion blending to iteratively update the latent motion using newly hypothesized motions, and a continuous space for motion is necessary for this motion blending. Specifically, we choose the required continuous space formed by the Cartesian product of trajectory in Euclidean space, root orientation in 6D continuous space~\cite{zhou2019continuity} and pose feature in SMPL-X prior space~\cite{pavlakos2019expressive}. For initialization, we directly set the latent motion as $\ddot{\mathcal{S}}_0=\hat{\mathcal{S}}_0$. Here, we first consider the latent route update. Given previous latent route $\ddot{\mathcal{R}}_{T-1}$ and current predicted route $\hat{\mathcal{R}}_{T}$, we can update the latent route as shown in Figure~\ref{fig:framework} (e) through blending for motion continuity:
\begin{equation}
    \label{eq:blending}
    \ddot{\mathcal{R}}_{T}[\tau] =  (1-\alpha^{\nu})\hat{\mathcal{R}}_{T}[\tau] + \alpha^{\nu}\ddot{\mathcal{R}}_{T-1}[\tau+1]
\end{equation}
where $\tau\in[0,\cdots,k-1]$, $\alpha = (k-\tau-1)/k$ is the time-variant blending coefficient, and $\nu$ is a hyper-parameter that equals to $4$ to make sure the responsiveness in our implementation. For $\ddot{\mathcal{R}}_{T-1}[k]$ which works as a placeholder in the formulation will be set to a zero vector. Inspired by previous works~\cite{hassan2021stochastic,wang2022towards} which utilize the $A^*$ algorithm to help path planning, we take the trajectory $\mathcal{T}^*$ given by the $A^*$ algorithm to adjust our predicted route. Unlike directly blending the route, we find that the low-frequency component of velocity given by $\mathcal{T}^*$ can help avoid obstacles. Therefore, we first calculate the velocity sequences $v^*$ and $\ddot{v}$ from $\mathcal{T}^*$ and $\ddot{\mathcal{T}}_T$ (trajectory of $\ddot{\mathcal{R}}_T$). Then, we utilize the Fast Fourier Transform to obtain the velocity spectrum:
\begin{equation}
    f^*=FFT(v^*)\hspace{0.3cm} and \hspace{0.3cm}\ddot{f}=FFT(\ddot{v}).
\end{equation}
After that, we blend the velocity sequence into the spectrum space through $\bar{f}=lf^* + (1-l)\ddot{f}$ where $l$ is a low-pass filter. The desired trajectory $\bar{\mathcal{T}}$ is derived from the blended velocity $\bar{v}=Real(IFFT(\bar{f}))$ via
\begin{equation}
    \bar{\mathcal{T}}(\tau) = \ddot{\mathcal{T}}_{T-1}[0] + \sum_{x=0}^{\tau}\bar{v}[x].
\end{equation}
Here, we can calculate the z-axis angular difference $M_{rot}$ between current trajectory and desired trajectory at the $m$-th frame in the future where $m$ is set to $15$. Then, the fused route can be adjusted in the matrix representation $\ddot{\mathcal{R}}_T = M_{rot}\ddot{\mathcal{R}}_T$. 
As for the poses, the latent pose sequence $\ddot{\mathcal{P}}_T$ is updated the same as Eq.~\ref{eq:blending}, while the latent motion is updated as $\ddot{\mathcal{S}}_T=(\ddot{\mathcal{R}}_T,\ddot{\mathcal{P}}_T)$.

The final synthesized motion from $0$ to $F$ is extracted as $\tilde{\mathcal{S}}_{0:F}=[\ddot{\mathcal{S}}_{0}[0],\cdots,\ddot{\mathcal{S}}_{F}[0]]$. We will terminate the generation or sample the next goal when the distance between goal and current position is smaller than a tolerance $\epsilon$.

Thanks to the anchor-based state and phase prediction, we can introduce a no-skating operation for contacted parts during iterative latent motion updates. Take locomotion for instance, we can know which part of the body should be in contact with the scene after checking current state $\hat{s}_T$ and phase $\hat{p}_T$. We will calculate the mean position of anchored body part $\mathbf{x}_{T-1}$ from the execution frame of previous latent motion $\ddot{\mathcal{S}}_{T-1}[0]$, and derive the new position  $\mathbf{x}_T$ from current latent motion $\ddot{\mathcal{S}}_{T}[0]$. Then, the current latent motion will be corrected for no-skating through a translation of $\kappa(\mathbf{x}_{T-1}-\mathbf{x}_T)$, where $\kappa$ is set to $0.6$ in our setting.

\section{Experiments}
\label{sec:exp}

\subsection{Datasets}
\label{subsec:datasets}

\noindent\textbf{PROX.} We take a dataset of real human motion in 3D scenes PROX~\cite{hassan2019resolving} for performance evaluation. It provides motion sequences (30FPS) of different subjects in $12$ scanned scenes. All route and pose information is represented by SMPL-X~\cite{pavlakos2019expressive} parameters and obtained through a fitting algorithm. Following Wang \etal~\cite{wang2021synthesizing}, we also take the fitted parameters as pseudo-ground truth (P-GT) and take motion sequences in 8 scenes for training and data in 4 residual scenes for testing. Each sample contains current information and goal position \& orientation as well as the motion in the future 2 seconds.

\noindent\textbf{GTA-IM.} We also take a dataset containing GTA motions in 3D scenes GTA-IM~\cite{cao2020long} for motion synthesis in more open spaces. It contains $102$ motion sequences (30FPS) of various characters in $10$ large scenes represented by RGB-D images. Each pose is represented by the location of $21$ skeleton joints. 

In order to align the data format of two datasets and unify the evaluation, we extend MMHuman3D~\cite{mmhuman3d} to fit the SMPL-X parameters using skeleton joints from GTA-IM where SMPL-X parameters of consecutive frames are constrained to ensure continuity, and we take these fitted parameters as pseudo-ground truth (P-GT). We also build a pipeline based on the Open3D~\cite{zhou2018open3d} to obtain the scene mesh, point cloud, as well as Signed Distance Function (SDF) of each scene from the RGB-D images. We refer to the training/testing split given by Wang \etal~\cite{wang2021scene} where $70$ sequences in $6$ scenes are used for training and $30$ sequences in $3$ scenes are used for testing. Similar to the data format of PROX, each sample contains current information, goal position\&orientation, and motion in the future 2 seconds, and all starting frames are sampled with a stride of 5 frames.

\subsection{Implementation Details}
\label{subsec:implement}
For detailed parameter settings, the spherical angular depth is fed into a small network consisting of several layers of light-weight spherical convolution and fully connected layers to extract the spherical angular pattern $f_a\in\mathbb{R}^{256}$. On the other side, the BEV elevation feature $f_b\in\mathbb{R}^{128}$ is extracted from the BEV elevation map through a small 2D convolution network.

In the GoalNet, the dimension of the goal condition feature $f_c^g$ is set to $128$, and the mean and variance value of the goal latent code $z^g$ is set to $\mu^g,(\sigma^g)^2\in\mathbb{R}^{32}$. The start pose feature $f_{sp}\in\mathbb{R}^{56}$ from the SMPL-X prior and the goal-based moving feature $f_g\in\mathbb{R}^{256}$ are used in the RouteNet. As for the PoseNet, there are additional route feature $f_r\in\mathbb{R}^{64}$ to supply route hints, while the dimension of pose sequence latent code $z^p$ is set to $64$.

In the training procedure, $\lambda_{1,\cdots,7}$ are set to $\{0.3$, $0.7$, $10$, $10$, $1$, $1$, $1\}$, respectively, for both PROX and GTA-IM datasets. GoalNet, RouteNet, and PoseNet are trained independently by Adam~\cite{kingma2014adam} optimizers with learning rates \{$10^{-3}$, $10^{-3}$, $3\times 10^{-6}$\}. All these networks are trained for $20$ epochs. 

Based on these settings, all our networks can be trained on a machine with 16G RAM and 11G VRAM, and the total networks' inference time is $4.9$ms on a single GTX 3090 after training.

\subsection{Evaluation Metrics}
\label{subsec:metric}

\noindent\textbf{Reconstruction and Physical Metric.} In order to evaluate how well different methods can reconstruct the future motion given the current subject information and goal position\&orientation as well as surrounding scenes, we follow Wang \etal~\cite{wang2021synthesizing} to calculate the mean L1 distance in motion translation, orientation, and pose parameter in SMPL-X latent space between the pseudo-ground truth and reconstructed motions. We also choose to take the Mean Per Joint Position Error (MPJPE) and Mean Per Vertex Position Error (MPVPE) for measurement as previous works~\cite{wang2021synthesizing,xu2021exploring} which are more corresponding to the visual difference. For the SMPL-X human meshes of predicted motion, we also consider physical metrics like the Contact and Non-collision scores~\cite{zhang2020psi,hassan2021populating} to check the ratio of bodies in contact with the scene and mesh vertices in free space.

\begin{figure}
    \centering
    \includegraphics[width=\linewidth]{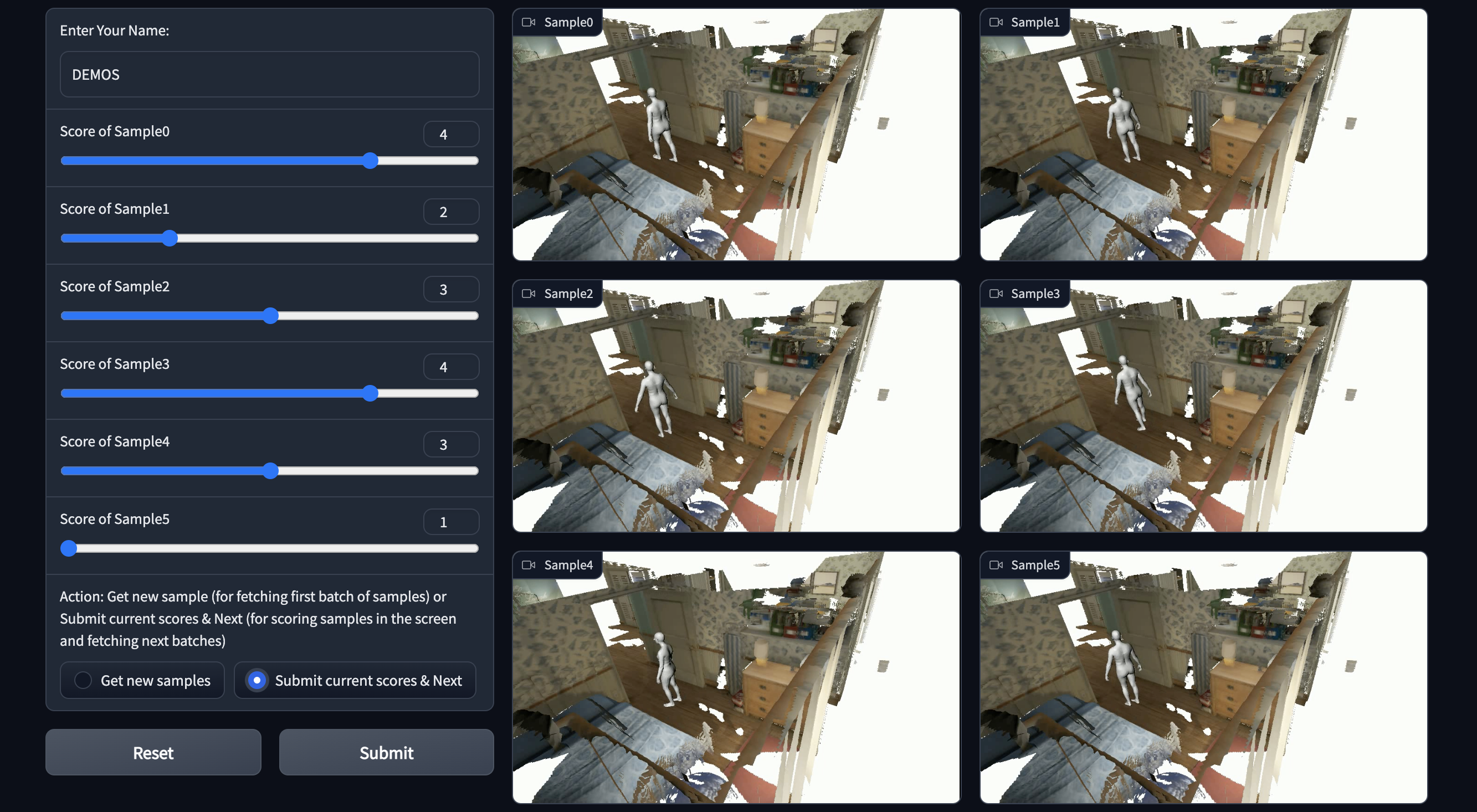}
    \caption{Overview of our online UI for the user study. (a) Users are asked to enter their username before fetching the video samples in case of repeated ratings. (b) Video samples are randomly permuted and presented to users. (c) Users can rate all samples after watching these videos. Meanwhile, they can terminate or continue this procedure at any time, and only submitted scores will be preserved.}
    \label{fig:ui}
\end{figure}

\noindent\textbf{Naturalness and Perception Metric.} In order to check the naturalness of the synthesized motion sequence, we additionally conduct a user study to rank the naturalness of generated motion given by different methods along with the pseudo-ground truth. For easier user study, we render all motions in 3D scenes into videos and design an online UI based on the Gradio framework~\cite{abid2019gradio} as shown in Figure~\ref{fig:ui}. 

\begin{figure*}[htb]
    \centering
    \includegraphics[width=\linewidth]{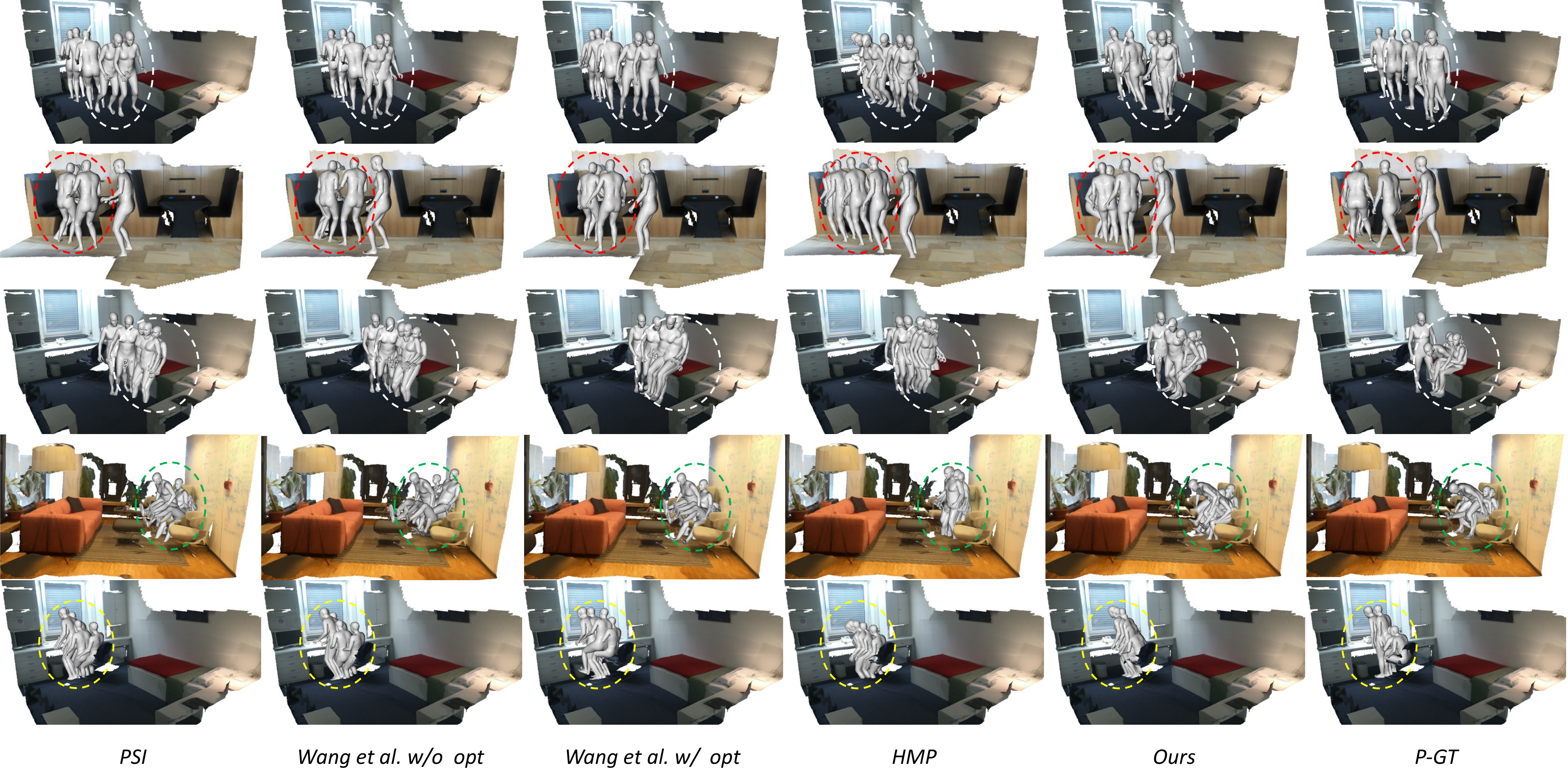}
    \caption{Visual comparison of different methods and pseudo-ground truth on predicting the motion in future 4 seconds given the same start human bodies and scanned scenes from PROX dataset.}
    \label{fig:prox_4s}
\end{figure*}

Users are required to enter their username to fetch the first batch of samples given by different methods/P-GT or continue the rating procedure. These samples with the same initial information are randomly permuted and then presented to users. They are asked to give a score between $1$ to $5$ for each sample after watching all videos within the batch. After submitting their scores, they can get a new batch of samples if they have not rated all motion samples. They can also terminate the rating at any time because we will record the results once they submit the scores and usernames can be used to continue the procedure. Finally, we will collect all the results and calculate the mean scores and variance values of different methods for both PROX~\cite{hassan2019resolving} and GTA-IM~\cite{cao2020long}.

In addition, we utilize Fr\'echet Distance (FD) as a perception metric like previous works~\cite{hassan2021stochastic,wang2022towards} to evaluate the perception similarity between the synthesized motions and P-GT motions in motion feature space.

\subsection{Experimental Results}
\label{subsec:results}

\noindent\textbf{Comparison on PROX.} The results of compared methods for motion reconstruction in the PROX~\cite{hassan2019resolving} dataset are reported in Table~\ref{tab:prox} where our method outperforms all previous works in the precision of position, orientation, and pose estimation. The contact and non-collision scores of predicted future motions are also reported in this table. We can see that our method has comparable or even better performance than previous methods in these two metrics even without an additional optimization stage. The results given by Wang \etal~\cite{wang2021synthesizing} with the optimization stage achieve the highest non-collision score, however, it can only be a trade-off between contact and non-collision in this situation and will be quite time-consuming for scene-aware motion synthesis.

\begin{table}[htb]
\setlength\tabcolsep{1.5pt}
\caption{Results of reconstruction and physical metric comparison between different competitors on PROX. }
\begin{footnotesize}
\begin{center}
\begin{tabular}{l|ccccccc}
\toprule
Methods & Transl. & Orient. & Pose & MPJPE & MPVPE & Cont. & Non-col.\\
\midrule
Wang \etal \cite{wang2021synthesizing} & 11.94& 12.88& 41.94& 33.39& 31.40& 99.84 & 93.88 \\
\ \  w/ opt \cite{wang2021synthesizing} & 14.06& 14.20& 57.13& 35.66& 34.18& 97.17 & \textbf{99.75} \\
Route+PSI \cite{zhang2020psi,wang2021synthesizing} & 11.94 & 12.88 & 38.21 & 32.08 & 30.66 & \textbf{99.92} & 93.06\\
HMP \cite{xu2021exploring} & 11.33& 15.97& 48.35& 30.83& 29.52& 99.85 & 94.49\\
Ours & \textbf{9.00}& \textbf{11.96}& \textbf{35.98}& \textbf{26.35}& \textbf{24.75}& 99.53 & 95.06\\
\bottomrule
\end{tabular}
\end{center}
\end{footnotesize}
\label{tab:prox} 
\end{table}

We also present the visual results of previous works and the proposed method in Figure~\ref{fig:prox_4s}. As can be seen in the first row, the turning around during locomotion given by our method is more continuous than the competitors. Meanwhile, the proposed method achieves a higher level of naturalness in the transition between locomotion and sitting, as indicated in the second row. In addition, our predicted sitting motions in the following three rows are more similar to the pseudo-ground truth and more compatible with surrounding scenes thanks to the specifically designed local scene perception. The generated sitting poses can change continuously according to the distance to the chair, and this is attributed to the geometry perception from the PBLSP module which guides the PoseNet to control scene-aware state coefficients and finish the sitting procedure.

\begin{figure*}
    \centering
    \includegraphics[width=\linewidth]{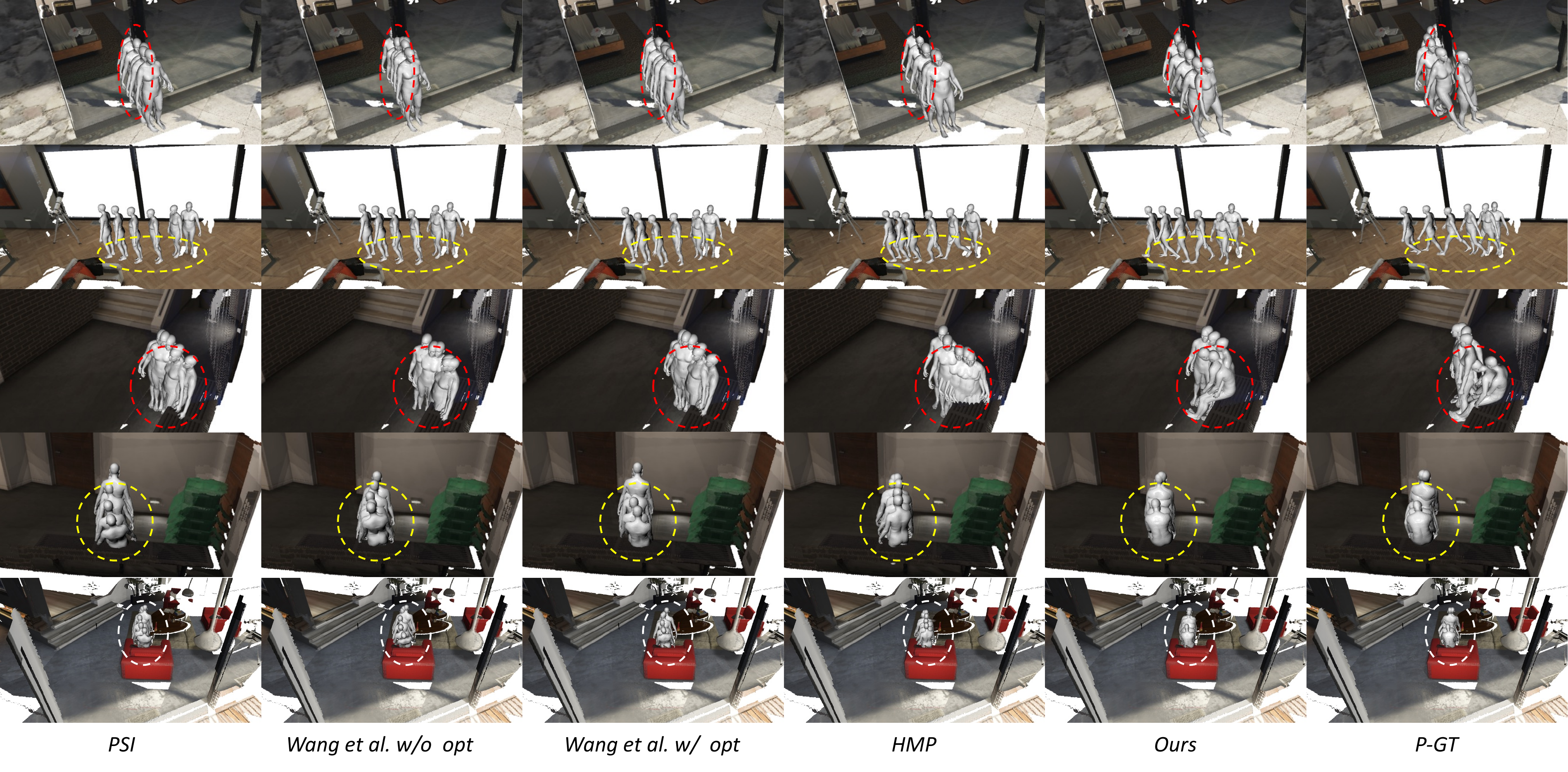}
    \caption{Visualization results of compared methods along with pseudo-ground truth on future motion prediction in GTA-IM.}
    \label{fig:gtaim_4s}
\end{figure*}

\noindent\textbf{Comparison on GTA-IM.} Here, we also report the motion reconstruction performance of different methods on GTA-IM dataset in Table~\ref{tab:gtaim}. This table shows that the proposed method can also achieve the best motion reconstruction performance among all compared methods. As for the physical metrics, synthesized motions predicted by our method achieve the highest non-collision score. Even though the optimization stage can improve the contact score of Wang \etal~\cite{wang2021synthesizing} on this dataset, it sacrifices the non-collision metric and results in a much lower score.

\begin{table}[htb]
\setlength\tabcolsep{1.5pt}
\caption{Results of prediction accuracy and physical plausibility comparison on GTA-IM.}
\begin{footnotesize}
\begin{center}
\begin{tabular}{l|ccccccc}
\toprule
Methods & Transl. & Orient. & Pose & MPJPE & MPVPE & Cont. & Non-col.\\
\midrule
Wang \etal \cite{wang2021synthesizing} & 5.73& 8.44& 69.54& 19.95& 17.74& 95.88 & 94.66 \\
\ \  w/ opt \cite{wang2021synthesizing} & 8.40& 8.71& 96.83& 24.85& 22.59& \textbf{99.35} & 90.93 \\
Route+PSI \cite{zhang2020psi,wang2021synthesizing} & 5.73 & 8.44 & 69.09 & 19.43 & 17.43 & 95.73 & 94.75 \\
HMP \cite{xu2021exploring} & 7.28& 9.91 & 78.98& 21.65& 19.67& 97.00 & 94.63\\
Ours & \textbf{4.88}& \textbf{7.38}& \textbf{54.28}& \textbf{15.28}& \textbf{13.86}& 97.89 & \textbf{95.82}\\
\bottomrule
\end{tabular}
\end{center}
\end{footnotesize}
\label{tab:gtaim} 
\end{table}

The visual comparison of different methods on GTA-IM is presented in Figure~\ref{fig:gtaim_4s}. Given the scene point cloud and initial human body information, all methods generate 4-second future motions with the highest probability. From the first row of Figure~\ref{fig:gtaim_4s}, we can see that the motion predicted by our method is more likely to circumvent the column compared to other methods due to the perception of obstacles. According to the second row, the phase information encoded by the discrete cosine transform utilized by HMP~\cite{xu2021exploring} can only provide limited phase hints for short-term motion, while the phase step information predicted by our method can better guide the locomotion and make sure the continuity for long-term motion synthesis. From the last few examples, we can see the proposed method can better reconstruct future motion with scene-aware actions given a goal and start human bodies in 3D scenes. That indicates the recognition of the local geometry in our method can help predict the correct anchor-based state to synthesize plausible motion.  

\noindent\textbf{Naturalness and Perception Similarity.} Human evaluations are taken into consideration to judge the naturalness of synthesized motions. For both the PROX and GTA-IM datasets, we generate $30$ samples for each method and present them to users for scoring. Finally, $224$ results are collected for each method ($3.73$ results per sample). We present the results of user study in Table~\ref{tab:naturalness}, and it demonstrates that the synthesized motions given by the proposed pipeline are more natural than those synthesized by previous methods. The average score of P-GT in GTA-IM is lower than that in PROX, and we think this is attributed to the synthetic essential of GTA-IM. 

\begin{table}[htb]
\setlength{\tabcolsep}{9.9pt}
\caption{Naturalness comparison based on user study [Score $\uparrow$]. The average score is reported along with the standard deviation in parentheses.
}
\begin{center}
\begin{small}
\begin{tabular}{l|c|c}
\hline
\multirow{2}{*}{Methods} & \multicolumn{2}{c}{Score $\uparrow$}\\
\cline{2-3}
%\cmidrule{2-6}
{}& \ \  PROX \ \  & GTA-IM\\
\hline
Wang \etal~\cite{wang2021synthesizing} & 1.62(0.61)& 1.64(0.75)\\
\hline
\ \  w/ opt \cite{wang2021synthesizing} & 2.33(0.71)& 1.86(0.74)\\
\hline
Route+PSI \cite{zhang2020psi,wang2021synthesizing} & 1.63(0.60) & 1.75(0.88)\\
\hline
HMP \cite{xu2021exploring} & 1.37(0.48)& 1.38(0.62)\\
\hline
Ours & 3.33(0.90) & 3.08(1.08) \\
\hline
P-GT & 4.50(0.86) & 3.45(1.14)\\
\hline
\end{tabular}
\end{small}
\end{center}
\label{tab:naturalness}
\end{table}

We also report the Fr\'echet Distance in motion feature space in Table~\ref{tab:fd} which indicates the perception similarity between synthesized motions and original ones. We can see that the synthesized motions given by our method are more similar to motions in the original datasets than other competitors.

\begin{table}[htb]
\setlength{\tabcolsep}{9.9pt}
\caption{Fr\'echet Distance [FD $\downarrow$] between synthesized motions and P-GT in motion feature distribution.
}
\begin{center}
\begin{small}
\begin{tabular}{l|c|c}
\hline
\multirow{2}{*}{Methods} & \multicolumn{2}{c}{FD[$\times10^2$] $\downarrow$} \\
\cline{2-3}
%\cmidrule{2-6}
{}& \ \ PROX\ \  & GTA-IM\\
\hline
Wang \etal~\cite{wang2021synthesizing} & 23.71 & 101.46\\
\hline
\ \  w/ opt \cite{wang2021synthesizing} & 28.87 & 110.10\\
\hline
Route+PSI \cite{zhang2020psi,wang2021synthesizing} & 19.33& 85.11 \\
\hline
HMP \cite{xu2021exploring} & 33.35&  110.83\\
\hline
Ours & 15.99 & 48.79 \\
\hline
\end{tabular}
\end{small}
\end{center}
\label{tab:fd}
\end{table}

\begin{figure*}[htb]
    \centering
    \includegraphics[width=\linewidth]{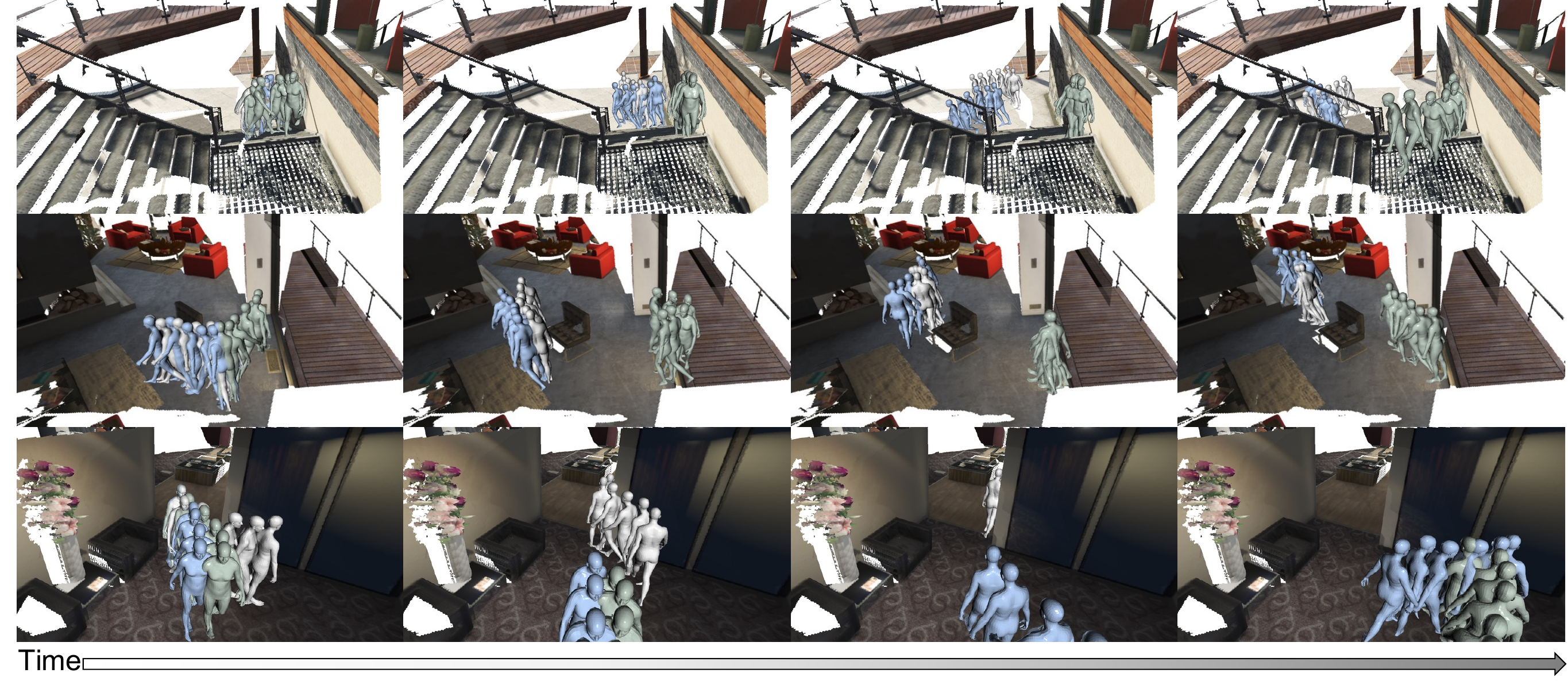}
    \caption{Visualization of different synthesized motions from the proposed pipeline given the same environment and initial situation.}
    \label{fig:diverse}
\end{figure*}

\noindent\textbf{Diversity in Motion Generation.} Due to the randomness in the sampling procedure of goal and pose generation, we can also benefit from the diversity in the synthesized human motions. Here, we show several synthesized motions in Figure~\ref{fig:diverse} that have the same initial human pose, position, and orientation, and were generated under the same surrounding environment. As can be seen, various future motions (shown in different colors) can be generated by our method given the same initial condition.

\begin{figure*}[htb]
    \centering
    \includegraphics[width=\linewidth]{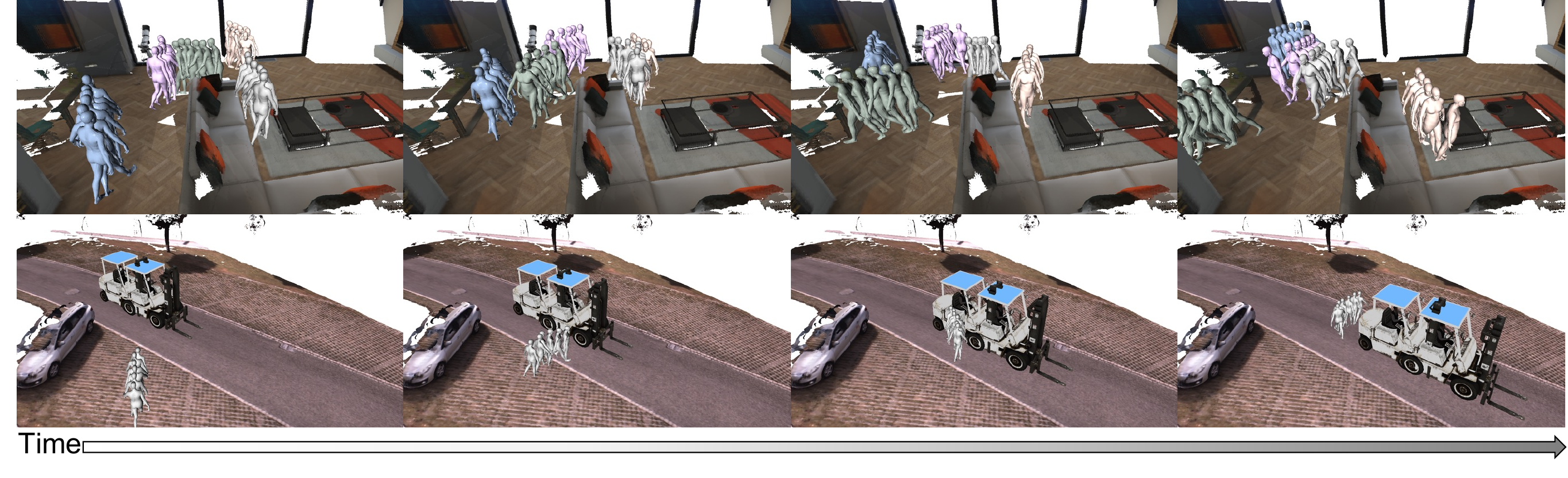}
    \caption{Visualization of motion synthesis in two representative dynamic environments. (a) Several persons are moving around in an indoor living room. (b) A fork is driving on the road in an outdoor scene.}
    \label{fig:dynamics}
\end{figure*}

\noindent\textbf{Motion Synthesis in Dynamic Environment.} In this paper, we also consider that the surrounding scene point cloud may change over time. The most common situation in the real world is that there are other persons or vehicles moving around in the scene. Here, we conduct experiments to evaluate the performance of the proposed methods in two typical dynamic environments: (1) several other persons moving around in an indoor scene from GTA-IM~\cite{cao2020long}, (2) a fork driving on the road in an outdoor scene from Semantic3D~\cite{hackel2017semantic3d}. The motions synthesized in these scenes are presented in Figure~\ref{fig:dynamics}. The results indicate our method can timely perceive their movement and update the latent motion to adapt to these changes in surrounding scene point clouds.

\subsection{Ablation Study}
\label{subsec:ablation}
In this section, we conducted more experiments to check the effectiveness of the proposed DEMOS pipeline and prove our claims. Without loss of generality, our ablation studies are mainly conducted on the GTA-IM dataset.

\begin{figure*}[htb]
    \centering
    \includegraphics[width=\linewidth]{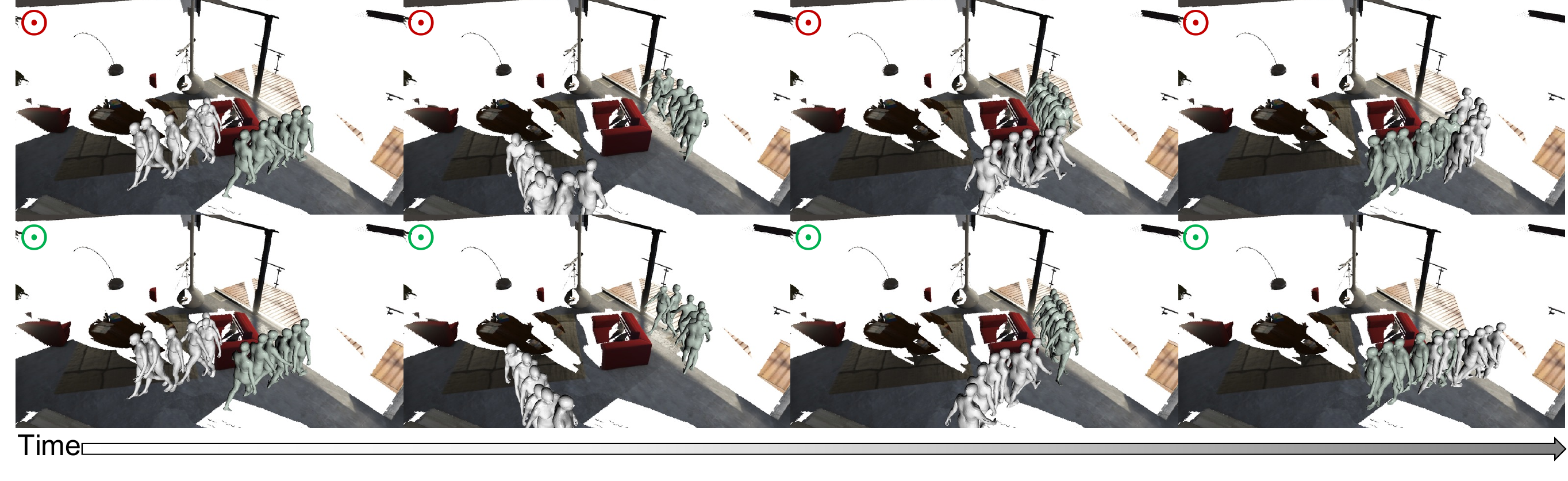}
    \caption{Comparison between the proposed iterative motion synthesis method {\color{odotred}$\odot$} and intuitive concatenation of synthesized motion sequences {\color{odotgreen}$\odot$}.}
    \label{fig:response}
\end{figure*}

\begin{figure*}[htb]
    \centering
    \includegraphics[width=\linewidth]{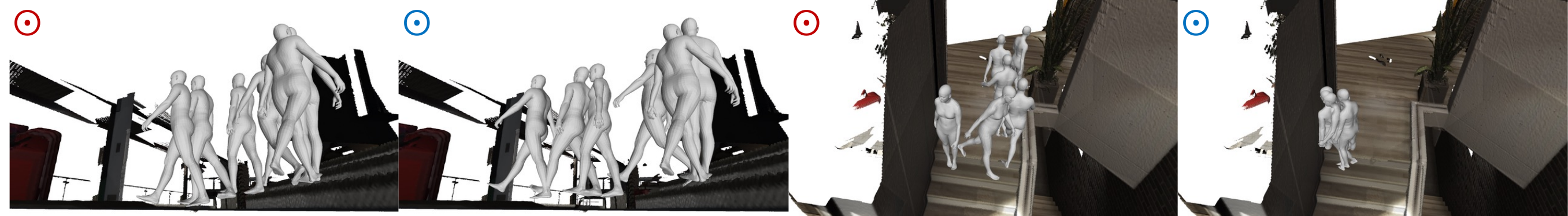}
    \caption{Two representative samples comparing the proposed iterative motion fusion {\color{odotred}$\odot$} with pure iterative pose generation method {\color{odotblue}$\odot$}.}
    \label{fig:stability}
\end{figure*}

\noindent\textbf{Instant Motion Prediction.} To evaluate the contribution of different components in our method to instant motion prediction, we first checked the performance of the pure backbone (annotated as Backbone). For the Backbone, we took the same network structures but without supervision on state and phase and only utilized PointNet to extract scene features. Then, we add the supervision given by the automatic annotation of Anchor-Based State (ABS) and the corresponding phase for cyclic locomotion. For the body-correlated Projection-based Local Scene Perception which exploits the relationship between the scene's local geometry and the start human body, we integrate the Spherical Angular Perception (SAP) and BEV Elevation Perception (BEP) into the PBLSP module gradually. 

We report all the results in Table~\ref{tab:ablation}. The results indicate that guiding the network to predict the state and phase of procedure can help synthesize motion with correct and unambiguous action. Moreover, recognizing the surrounding scene point cloud via Spherical-BEV perception and leveraging its correlation with human start information can enhance the precision of the predicted route and motion. Specifically, the SAP focuses more on route prediction and we think that is because SAP can better recognize the objects in different directions. On the other hand, the BEP is more effective in pose sequence prediction because the body-centered elevation recognition can further provide height hints of surrounding objects for state prediction. These functional components can help the networks to predict scene-aware motions instantly via better recognizing the surrounding environment and exploiting the correlation between the scene and the human body.

\begin{table}[htb]
\setlength\tabcolsep{0.9pt}
\caption{Ablation study on the impact of different components of the proposed method on motion prediction accuracy in GTA-IM. }
\begin{footnotesize}
\begin{center}
\begin{tabular}{l|ccccccc}
\toprule
Methods & Transl. & Orient. & Pose & MPJPE & MPVPE & Cont. & Non-col.\\
\midrule
Backbone & 5.64 & 8.62 & 70.06& 19.35 & 17.16 & 96.69 & 94.89\\
\ \ +ABS & 5.64& 8.62& 64.78& 17.49& 15.98& 95.30& 95.78\\
\ \ \ \ +SAP& 5.07& 7.46& 61.68& 16.26& 14.63& 96.70& \textbf{95.85}\\
\ \ \ \ \ \ +BEP& \textbf{4.88}& \textbf{7.38}& \textbf{54.28}& \textbf{15.28}& \textbf{13.86}& \textbf{97.89} & 95.82\\
\bottomrule
\end{tabular}
\end{center}
\end{footnotesize}
\label{tab:ablation} 
\end{table}

\noindent\textbf{Spherical Perception.} In the PBLSP module, the proposed Spherical Angular Perception (SAP) is utilized to recognize the obstacles or contact objects at different directions. Compared to common spherical convolutions, the network of SAP is designed in a more lightweight way to extract scene hints from the angular depth map. For comparison with common spherical convolution, we replace the network of SAP with KTN~\cite{su2019kernel} and reported the results in Table~\ref{tab:spherical}. The results indicate the proposed lightweight network of SAP can achieve comparable performance thanks to the pre-computed angular depth representation.

\begin{table}[htb]
\setlength\tabcolsep{0.9pt}
\caption{Ablation study of spherical perception on motion prediction accuracy in GTA-IM. }
\begin{footnotesize}
\begin{center}
\begin{tabular}{l|ccccccc}
\toprule
Methods & Transl. & Orient. & Pose & MPJPE & MPVPE & Cont. & Non-col.\\
\midrule
Ours(SAP)& 4.88& 7.38 & 54.28& 15.28& 13.86& 97.89 & 95.82\\
Ours(KTN)& 4.86& 7.20& 55.35& 15.22& 13.80& 96.76 & 96.05\\
\bottomrule
\end{tabular}
\end{center}
\end{footnotesize}
\label{tab:spherical} 
\end{table}

\noindent\textbf{Responsiveness.} In the proposed pipeline, we iteratively hypothesize new motion sequences with the help of motion prior from the auto-encoding scheme and utilize them to update latent motion, making the final synthesized motion both stable and responsive to the dynamic environment. In previous works, it is more common to generate the motion in the future $k$ frames $\hat{\mathcal{S}}_T$ given the current information, and then take the final frame $k$ frames later (sub-goal in Wang \etal~\cite{wang2021synthesizing}) as the new initial information for next motion synthesis. When generating the next hypothesized motions $\hat{\mathcal{S}}_{T+k}$, the surrounding scene point cloud can also be updated. Lastly, $\hat{\mathcal{S}}_{0},\hat{\mathcal{S}}_{k},\cdots,\hat{\mathcal{S}}_{F-k}$ will be concatenated as the final synthesized motion. Here, we attempt to compare these two strategies to prove the advantage of the proposed pipeline in responsiveness. Figure~\ref{fig:response} illustrates the results of two strategies when another person is moving around. The visualization results indicate that naive motion sequence generation and concatenation is less likely to circumvent another moving person and unable to accommodate dynamic scene point cloud.

\noindent\textbf{Stability.}
Furthermore, our method demonstrates better stability on long-term motion synthesis in scanned point cloud scenes when compared to the pure iterative competitor with an identical network structure. Here, we iteratively generate the body route and pose parameter $(\hat{R},\hat{P})_{T}$ in the next frame, and combine all of them to obtain the synthesized motion $(\hat{R},\hat{P})_{0:F}$. We visualize two typical samples of the compared methods in Figure~\ref{fig:stability}. The results indicate that the pure iterative method is more prone to diverge from a natural trajectory or collapse to a mean solution, which is unfavorable in the context of long-term motion synthesis.

\section{Conclusion}
In this paper, we design a Dynamic Environment Motion Synthesis (DEMOS) framework that benefits both from the stability of prior-based motion generation and the responsiveness of iterative motion updating. We introduce a Spherical-BEV perception to make the instantly generated motion consistent with surrounding geometry without the need for an optimization stage. Thus, we can iteratively blend new hypothesized motion with latent motion in an online manner, making the synthesized motion accommodate to the dynamic environment. We align the data format of PROX and GTA-IM to unify the training and evaluation procedure. Versatile experiments demonstrate the proposed method can generate more precise and natural motions which are also plausible in various dynamic scene point clouds. Even though, this task is still quite challenging and in the early stage where more properties like higher method integration and easier control of motion style can be pursued based on the current pipeline in future works.

% if have a single appendix:
%\appendix[Proof of the Zonklar Equations]
% or
%\appendix  % for no appendix heading
% do not use \section anymore after \appendix, only \section*
% is possibly needed

% use appendices with more than one appendix
% then use \section to start each appendix
% you must declare a \section before using any
% \subsection or using \label (\appendices by itself
% starts a section numbered zero.)
%

\iffalse
\appendices
\section{Proof of the First Zonklar Equation}
Appendix one text goes here.

% you can choose not to have a title for an appendix
% if you want by leaving the argument blank
\section{}
Appendix two text goes here.

% use section* for acknowledgment

\ifCLASSOPTIONcompsoc
  % The Computer Society usually uses the plural form
  \section*{Acknowledgments}
\else
  % regular IEEE prefers the singular form
  \section*{Acknowledgment}
\fi
The authors would like to thank...
\fi

% Can use something like this to put references on a page
% by themselves when using endfloat and the captionsoff option.
\ifCLASSOPTIONcaptionsoff
  \newpage
\fi

% trigger a \newpage just before the given reference
% number - used to balance the columns on the last page
% adjust value as needed - may need to be readjusted if
% the document is modified later
%\IEEEtriggeratref{8}
% The "triggered" command can be changed if desired:
%\IEEEtriggercmd{\enlargethispage{-5in}}

% references section

% can use a bibliography generated by BibTeX as a .bbl file
% BibTeX documentation can be easily obtained at:
% http://mirror.ctan.org/biblio/bibtex/contrib/doc/
% The IEEEtran BibTeX style support page is at:
% http://www.michaelshell.org/tex/ieeetran/bibtex/
\bibliographystyle{IEEEtran}
% argument is your BibTeX string definitions and bibliography database(s)
\bibliography{main}
%
% <OR> manually copy in the resultant .bbl file
% set second argument of \begin to the number of references
% (used to reserve space for the reference number labels box)
%\begin{thebibliography}{1}
%\bibitem{IEEEhowto:kopka}
%H.~Kopka and P.~W. Daly, \emph{A Guide to \LaTeX}, 3rd~ed.\hskip 1em plus
%  0.5em minus 0.4em\relax Harlow, England: Addison-Wesley, 1999.
%\end{thebibliography}

% biography section
% 
% If you have an EPS/PDF photo (graphicx package needed) extra braces are
% needed around the contents of the optional argument to biography to prevent
% the LaTeX parser from getting confused when it sees the complicated
% \includegraphics command within an optional argument. (You could create
% your own custom macro containing the \includegraphics command to make things
% simpler here.)
%\begin{IEEEbiography}[{\includegraphics[width=1in,height=1.25in,clip,keepaspectratio]{mshell}}]{Michael Shell}
% or if you just want to reserve a space for a photo:

\begin{IEEEbiography}[{\includegraphics[width=1in,height=1.25in,clip,keepaspectratio]{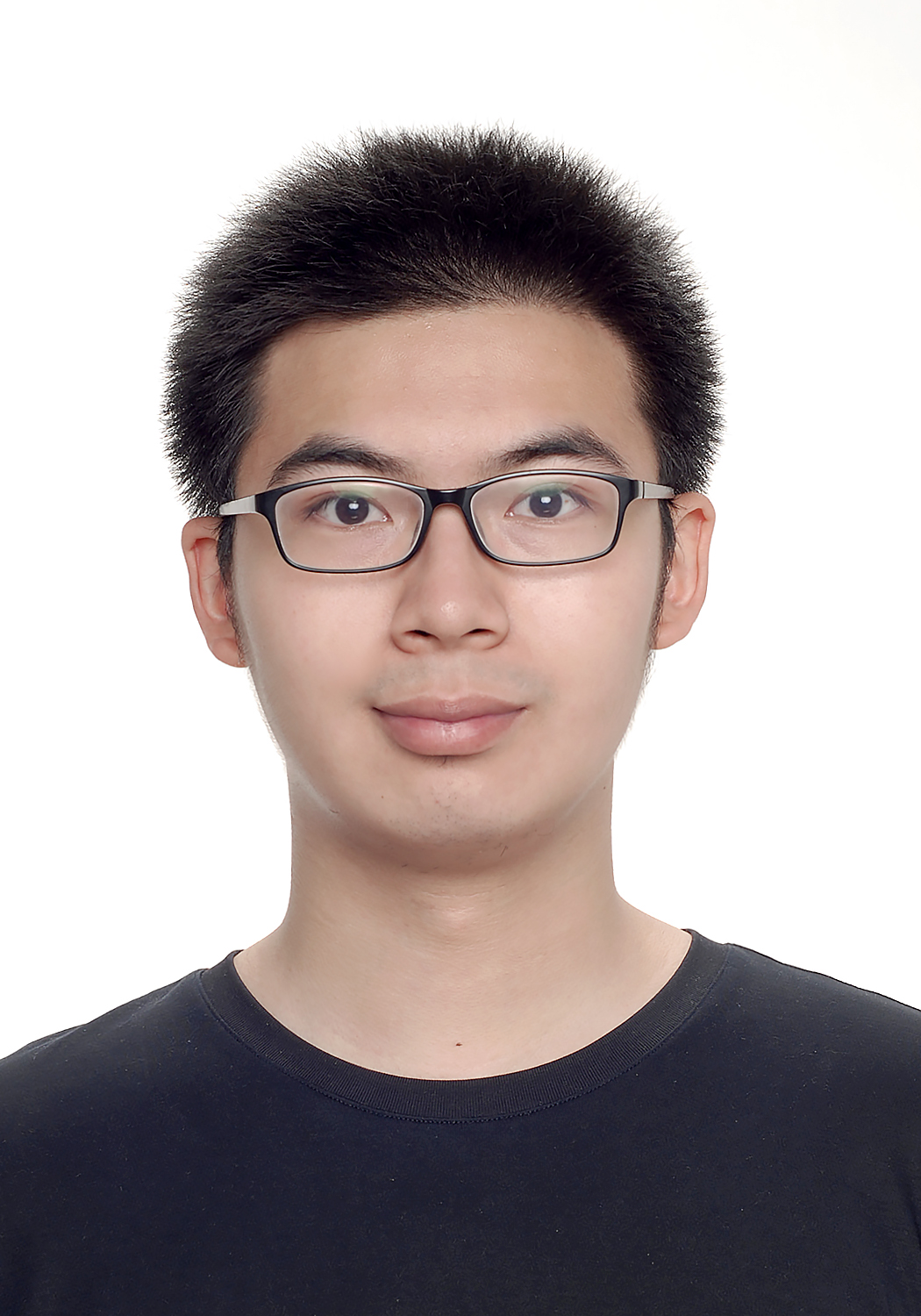}}]{Jingyu Gong} received his B.S. degree in Physics from the Shanghai Jiao Tong University, China in 2019. He is now a Ph.D. student at the Department of Computer Science and Engineering, Shanghai Jiao Tong University, China. His research interests cover 3D point cloud recognition and 3D motion synthesis.
\end{IEEEbiography}

\begin{IEEEbiography}[{\includegraphics[width=1in,height=1.25in,clip,keepaspectratio]{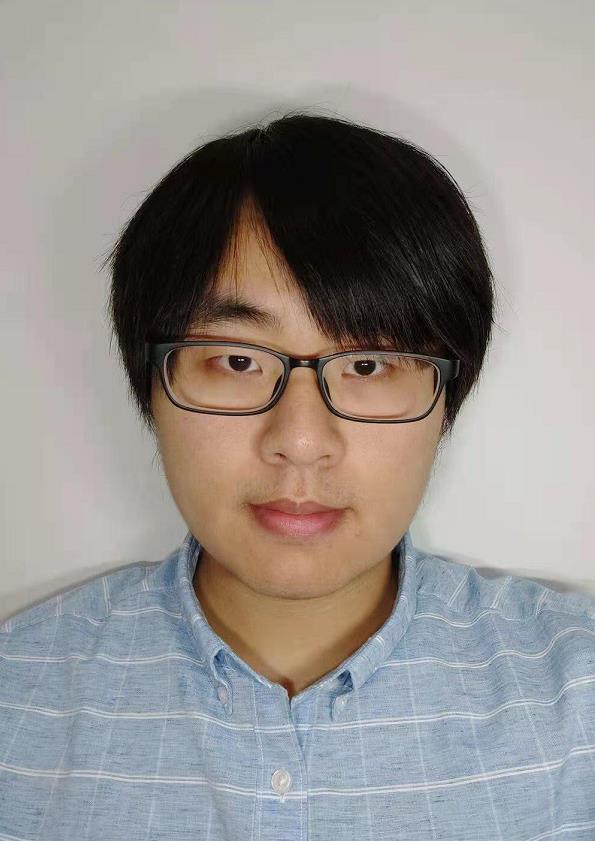}}]{Min Wang} received his Ph.D. degrees from the Department of Computer Science and Technologies, Shanghai Jiao Tong University. He is currently the senior researcher of SenseTime, responsible for the research of human-computer interaction and AI content generation. His research interest includes 3D Vision and Computer Graphics.
\end{IEEEbiography}

\begin{IEEEbiography}[{\includegraphics[width=1in,height=1.25in,clip,keepaspectratio]{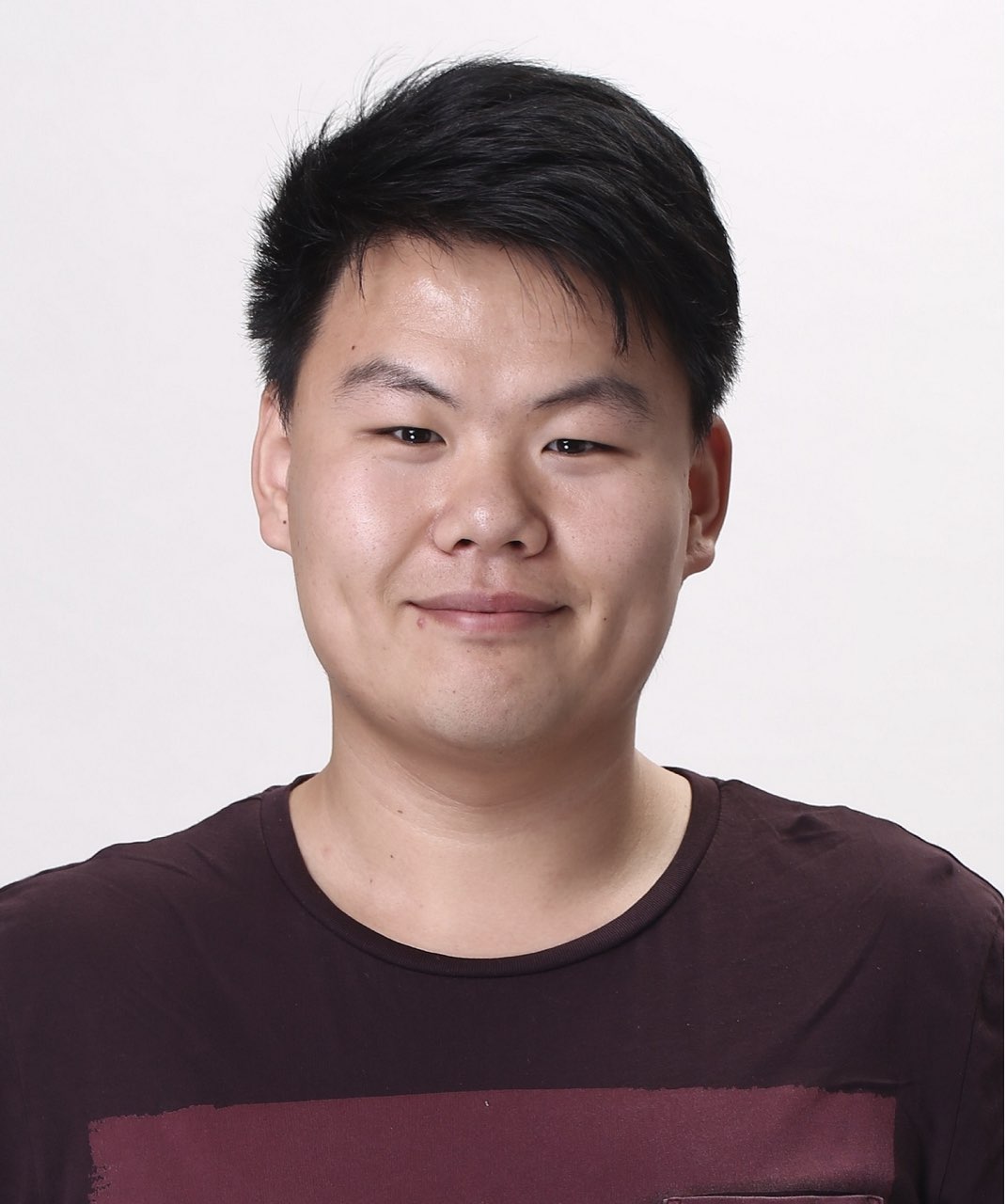}}]{Wentao Liu} received the PhD degree from the School of EECS, Peking University. He is currently the research director of SenseTime, responsible for end-edge computing research. The research products are widely applied in augmented reality, smart industry, and business intelligence. His research interests include computer vision and pattern recognition.
\end{IEEEbiography}

\begin{IEEEbiography}[{\includegraphics[width=1in,height=1.25in,clip,keepaspectratio]{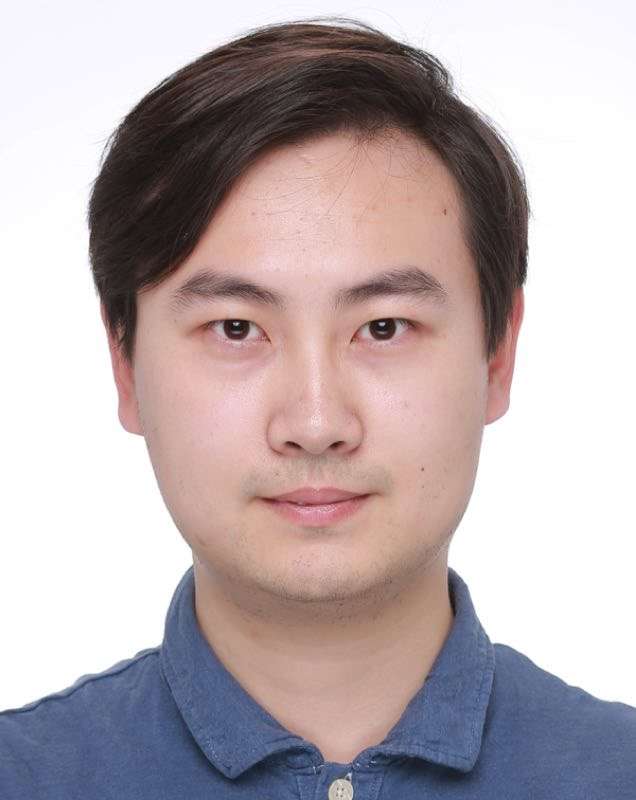}}]{Chen Qian} is currently the executive research director of SenseTime, where he is responsible for leading the team in AI content generation and end-edge computing research in 2D and 3D sce- narios. The technology is widely used in the top four mobile companies in China, APPs both home and abroad in augmented reality, video sharing and live streaming, vehicle OEMs, and smart industry. He has published dozens of articles on top journals and dozens of papers on top conferences, such as IEEE Transactions on Pattern Analysis and Machine Intelligence, CVPR, ICCV, and ECCV with more than 4000 citations. He has also led the team to achieve the first place in the Competition of Face Identification and Face Verification in Megaface Challenge.
\end{IEEEbiography}

\begin{IEEEbiography}[{\includegraphics[width=1in,height=1.25in,clip,keepaspectratio]{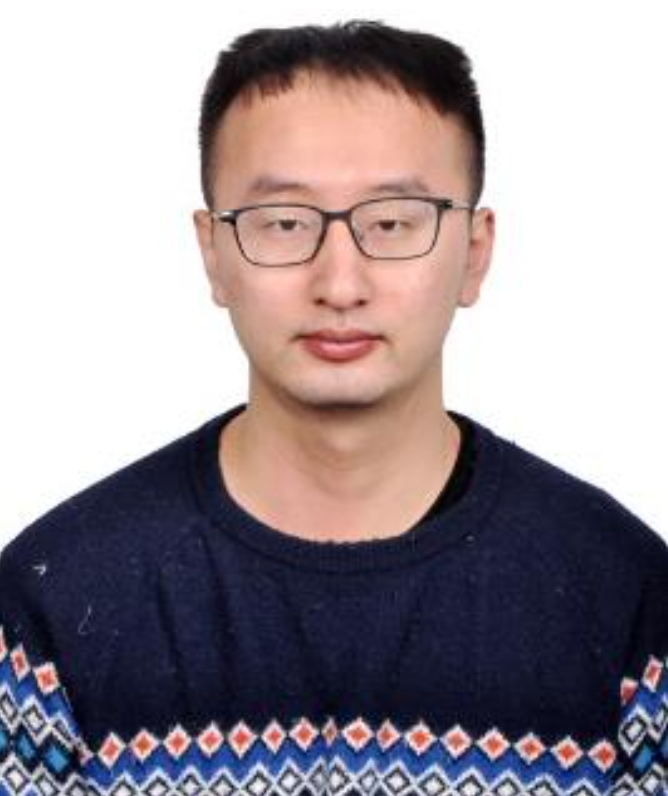}}]{Zhizhong Zhang} received the Ph.D. degree in pattern recognition and intelligent systems from the Institute of Automation, Chinese Academy of Sciences (CAS), in 2020. He is currently an Associate Research Professor with the School of Computer Science and Technology, East China Normal University. His research interests include image processing, computer vision, machine learning, and pattern recognition.
\end{IEEEbiography}

\begin{IEEEbiography}[{\includegraphics[width=1in,height=1.25in,clip,keepaspectratio]{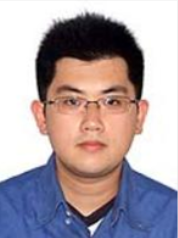}}]{Yuan Xie} received the PhD degree in Pattern Recognition and Intelligent Systems from the Institute of Automation, Chinese Academy of Sciences (CAS), in 2013. He is currently a full professor with the School of Computer Science and Technology, East China Normal University, Shanghai, China. His research interests include image processing, computer vision, machine learning, and pattern recognition. He has published around 90 papers in major international journals and conferences including the IJCV, IEEE TPAMI, TIP, TNNLS, TCYB, NIPS, ICML, CVPR, ECCV, ICCV, etc. He also has served as a reviewer for more than 15 journals and conferences. Dr. Xie received the National Science Fund for Excellent Young Scholars 2022.
\end{IEEEbiography}

\begin{IEEEbiography}[{\includegraphics[width=1in,height=1.25in,clip,keepaspectratio]{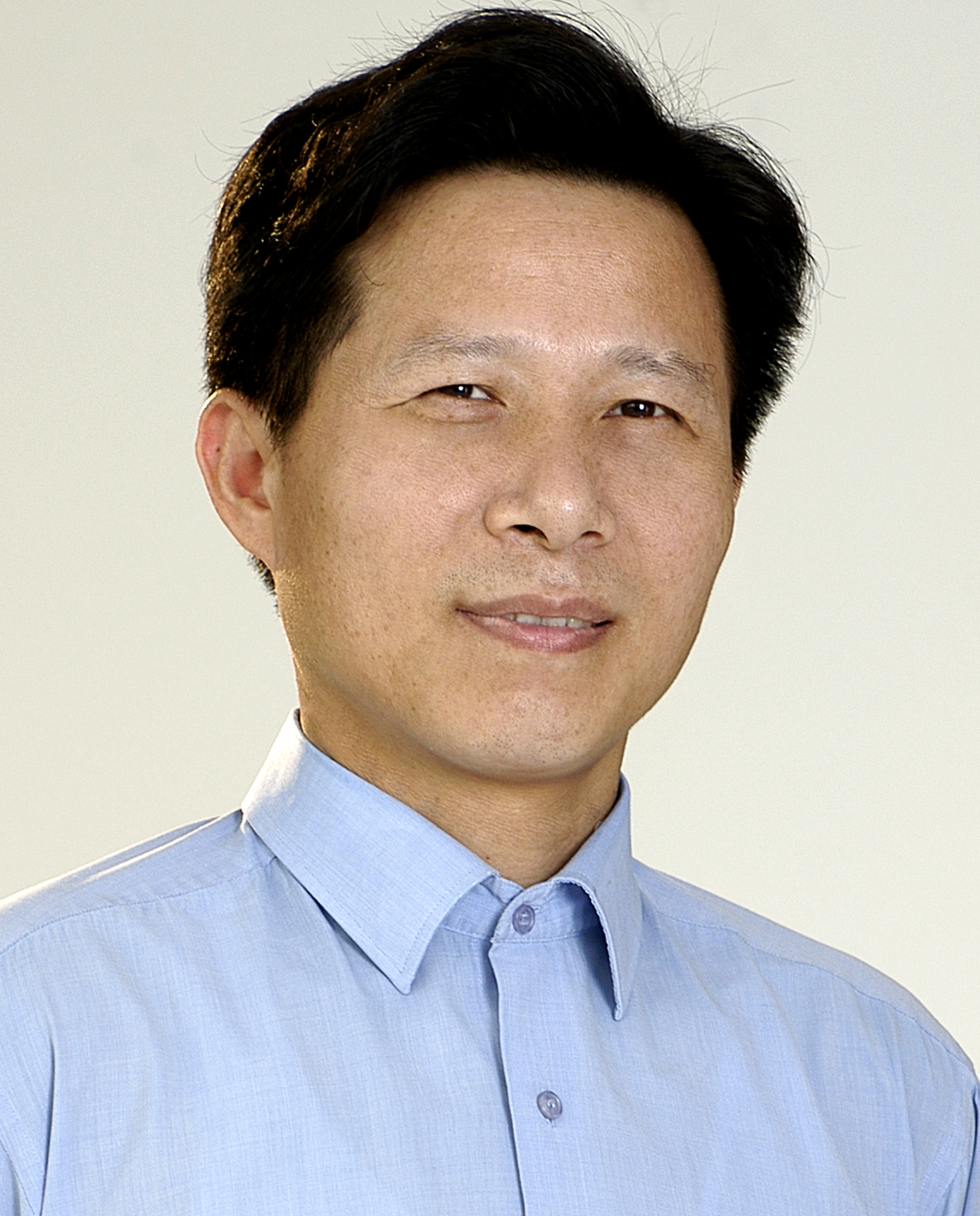}}]{Lizhuang Ma} received his B.S. and Ph.D. de- grees from the Zhejiang University, China in 1985 and 1991, respectively. He is now a Distinguished Professor, at the Department of Computer Science and Engineering, Shanghai Jiao Tong University, China and the School of Computer Science and Technology, East China Normal University, China. He was a Visiting Professor at the Frounhofer IGD, Darmstadt, Germany in 1998, and a Visiting Professor at the Center for Advanced Media Technology, Nanyang Technological University, Singapore from 1999 to 2000. His research interests include computer vision, computer aided geometric design, computer graphics, scientific data visualization, computer animation, digital media technology, and theory and applications for computer graphics, CAD/CAM. He serves as the reviewer of IEEE TPAMI, IEEE TIP, IEEE TMM, CVPR, AAAI etc.
\end{IEEEbiography}

% You can push biographies down or up by placing
% a \vfill before or after them. The appropriate
% use of \vfill depends on what kind of text is
% on the last page and whether or not the columns
% are being equalized.

%\vfill

% Can be used to pull up biographies so that the bottom of the last one
% is flush with the other column.
%\enlargethispage{-5in}

% that's all folks
\end{document}